\documentclass{article}

\usepackage{PRIMEarxiv}

\errorcontextlines\maxdimen

\usepackage[utf8]{inputenc} 
\usepackage[T1]{fontenc}    
\usepackage{url}            
\usepackage{booktabs}       
\usepackage{nicefrac}       
\usepackage{microtype}      
\usepackage{fancyhdr}       
\usepackage{graphicx}
\usepackage{xcolor}
\usepackage{threeparttable}
\usepackage{subcaption}

\usepackage{lineno,hyperref}
\hypersetup{colorlinks, citecolor=blue, linkcolor=red, urlcolor=green}
\usepackage{amsmath,amsfonts,amssymb}
\usepackage{graphics}
\usepackage{bm}
\usepackage{multirow}

\usepackage{algorithm}
\usepackage{algpseudocode}

\usepackage{mathtools}
\usepackage{siunitx}
\DeclarePairedDelimiterXPP\BigOSI[2]%
  {\mathcal{O}}{(}{)}{}%
  {\SI{#1}{#2}}

\title{A foundational neural operator that continuously learns without forgetting}

\author{Tapas Tripura\\
  Department of Applied Mechanics\\
  Indian Institute of Technology Delhi\\
  Hauz Khas, 110016, India\\
  \texttt{tapas.t@am.iitd.ac.in} \\
  \And
  Souvik Chakraborty \\
  Department of Applied Mechanics\\
  Yardi School of Artificial Intelligence (ScAI)\\
  Indian Institute of Technology Delhi\\
  Hauz Khas, 110016, India\\
  \texttt{souvik@am.iitd.ac.in} \\
}

\begin{document}
\maketitle

\begin{abstract}
Machine learning has witnessed substantial growth, leading to the development of advanced artificial intelligence models crafted to address a wide range of real-world challenges spanning various domains, such as computer vision, natural language processing, and scientific computing. Nevertheless, the creation of custom models for each new task remains a resource-intensive undertaking, demanding considerable computational time and memory resources. In this study, we introduce the concept of the Neural Combinatorial Wavelet Neural Operator (NCWNO) as a foundational model for scientific computing. This model is specifically designed to excel in learning from a diverse spectrum of physics and continuously adapt to the solution operators associated with parametric partial differential equations (PDEs). The NCWNO leverages a gated structure that employs local wavelet experts to acquire shared features across multiple physical systems, complemented by a memory-based ensembling approach among these local wavelet experts. This combination enables rapid adaptation to new challenges. The proposed foundational model offers two key advantages: (i) it can simultaneously learn solution operators for multiple parametric PDEs, and (ii) it can swiftly generalize to new parametric PDEs with minimal fine-tuning. The proposed NCWNO is the first foundational operator learning algorithm distinguished by its (i) robustness against catastrophic forgetting, (ii) the maintenance of positive transfer for new parametric PDEs, and (iii) the facilitation of knowledge transfer across dissimilar tasks. Through an extensive set of benchmark examples, we demonstrate that the NCWNO can outperform task-specific baseline operator learning frameworks with minimal hyperparameter tuning at the prediction stage. We also show that with minimal fine-tuning, the NCWNO performs accurate combinatorial learning of new parametric PDEs.
\end{abstract}


\section{Introduction}
Task-specific machine learning algorithms are currently undergoing a major shift with a machine learning model trained on a broad set of general tasks and then adapting the model for a wide set of downstream tasks. This class of representation learning algorithms is referred to as the foundation models, a term first coined by Stanford's Centre for Human-Centered Artificial Intelligence \cite{bommasani2021opportunities}. Foundation models employ modular learning \cite{veness2021gated} and transfer learning \cite{thrun1998lifelong} to transfer the previous experience to new tasks, thereby improving the resource efficiency of task-specific learning. Existing examples of foundation models are BERT \cite{kenton2019bert}, GPT3 \cite{brown2020language}, CLIP \cite{radford2021learning}, ALIGN \cite{jia2021scaling}, and PALM \cite{chowdhery2022palm}. Despite the recent developments, foundation models are mostly limited to computer vision \cite{yuan2023power}, large language modeling \cite{zhou2023comprehensive}, and information retrieval \cite{guu2020retrieval}. 
While all the concepts of data efficiency, predictive accuracy, and generalization play a key role in scientific computing, no effort has been made to develop foundation models for scientific computing. In this work, we propose the first foundation model for accelerating complex simulations in scientific computing that can be trained on datasets from a broad set of different and complex partial differential equations (PDEs) and fine-tuned later to adapt the trained model for new unseen tasks. 

Natural systems are governed by conservation and constitutive laws that are scientifically modeled as complex PDEs. The study of underlying scientific processes driving the natural systems in science and engineering often involves repeated simulation of complex PDEs for different input parameters, such as different system parameters, different initial/boundary conditions (ICs/BCs), different geometry conditions, and different source functions. In the absence of tractable analytical tools, solving these so-called parametric PDEs (class of PDEs whose parameters are allowed to vary over a finite range) requires repetitive execution of resource-intensive numerical techniques such as finite/spectral element methods (FEM/SEM) \cite{hughes2012finite, lord2014introduction} and finite volume methods (FVM) \cite{moukalled2016finite}. This entails initiating separate calculations from scratch for each combination of input parameters.
With a balance between the computational cost and reduced accuracy cum generalization to new unseen input parameters, reduced-order methods are able to reduce the computational complexity by learning surrogate models from existing datasets \cite{lucia2004reduced,peherstorfer2016data,lassila2014model,majda2018strategies}. Contemporary developments in neural networks have also contributed to the development of fast emulators for accelerated scientific computing of complex PDEs \cite{raissi2019physics,sun2020surrogate,zhu2019physics,sirignano2018dgm,goswami2020transfer,chakraborty2021transfer}. In an attempt to strike a balance between computational expense and the ability to generalize to previously unseen input parameters, reduced-order methods have emerged as a valuable approach. These methods achieve a reduction in computational complexity by constructing surrogate models based on existing datasets \cite{lucia2004reduced, peherstorfer2016data, lassila2014model, majda2018strategies}. Furthermore, contemporary advancements in neural networks have played a significant role in the development of swift emulators designed for the accelerated computational analysis of complex PDEs \cite{raissi2019physics, sun2020surrogate, zhu2019physics, sirignano2018dgm, goswami2020transfer, chakraborty2021transfer}.
Although reduced-order models and neural network-based approaches have succeeded in mitigating the time complexity to a certain extent by learning rapid emulators from available data, there remains an inescapable requirement for independent model training from scratch when dealing with each new combination of input parameters. 

A relatively new paradigm referred to as neural operators has emerged in recent years \cite{li2020fourier,li2020neural,gupta2021multiwavelet,lu2021learning,wang2021learning,tripura2023wavelet,tripura2023physics}. Neural operators remedy the need for repeated training for each input case by learning the underlying solution operator (map between infinite-dimensional function spaces) of parametric PDEs as opposed to the artificial neural networks (ANN) that learn the map between two finite-dimensional vector spaces. Pioneering works in the neural operator include the graph neural operator (GNO) \cite{li2020neural}, the Fourier neural operator (FNO) \cite{li2020fourier}, and the deep operator network (DeepONet) \cite{lu2021learning}. While the GNO and FNO are motivated by the Greens kernel function from classical functional analysis \cite{yosida2012functional}, the DeepONet and its Bayesian alternative \cite{garg2023vb} use the universal approximation theorem for operators \cite{chen1995universal,back2002universal}. Learning operator kernels using autoencoder in Koopman coordinates is also demonstrated in \cite{navaneeth2022koopman}. 
The learning of operator kernels in the wavelet space is evident in the wavelet neural operator (WNO), which is proposed for solving parametric PDEs in computational mechanics \cite{tripura2023wavelet} and medical elastography \cite{tripura2023elastography}. Motivated by the physics-informed neural network, the physics-informed DeepONet \cite{wang2021learning}, physics-informed FNO \cite{li2021physics}, and physics-informed WNO \cite{tripura2023physics} were also proposed by biasing the network output functions to satisfy the physics of underlying PDE to achieve enhanced generalization and 100\% data efficiency. Non-neural network approaches like the operator-valued kernel method are also proposed in \cite{kadri2016operator,griebel2017reproducing}. 

While a growing number of interests can be observed in developing and applying neural operators for solving all sorts of real-world problems, existing neural operators can learn the solution operators of only a single parametric PDE and can not adapt to new PDE systems without catastrophic forgetting. 
Motivated by the contemporary developments in modular \cite{veness2021gated} and combinatorial learning \cite{wang2020combinatorial}, we propose the neural combinatorial wavelet neural operator (NCWNO) that can be trained to learn the universal solution operator of a broad set of different parametric PDEs and later deployed as a foundation model to learn solution operators of new parametric PDEs by fine-tuning on a relatively small dataset for relatively small training epochs. 
The proposed NCWNO is composed of three major components: (i) expert wavelet kernel integral blocks consisting of an ensemble of local wavelet experts modeled as wavelet kernel integral \cite{tripura2023wavelet}, (ii) local wavelet integral blocks for parameterizing the local wavelet experts, and (iii) a gating mechanism for task-specific channel-level mixing of predictions from local experts. Combined with the local experts, the gating mechanism allows the combinatorial transfer of knowledge from the previously learned universal solution operator to new parametric PDEs without catastrophic forgetting.  
Unlike the architecture-based modeling of experts, where the model averaging is performed between models of different architecture, we instantiate the experts locally within each expert wavelet kernel integral block of the NCWNO. Each local wavelet expert uses different wavelet basis functions within the same model architecture to parameterize the local experts in a different wavelet domain. The gating mechanism is designed as a function of the actual parametric inputs and the corresponding PDE label, outputting the local probabilities of each local wavelet expert. 

The salient features of the proposed NCWNO can be encapsulated into the following points: (i) \textbf{Learning universal operator}: The NCWNO learns the most generalized solution operator of the underlying set of parametric PDEs by performing channel-level averaging of local wavelet experts in-a-go. Therefore, the NCWNO can approximate a broad set of physical systems during inference by varying the model probabilities without further fine-tuning (achieving zero-shot prediction).
(ii) \textbf{Increased accuracy on multiple tasks}: While a PDE-specific neural operator outperforms other models for a specific PDE, it does not generalize to other parametric PDEs. The proposed NCWNO solves this problem by exploiting the locally different behaviors of each local expert.
(iii) \textbf{Minimal tuning}: The ensembling of experts in the NCWNO is performed locally at the channel level instead of a mixture of models of different architecture. Therefore, the cost of parameter-tuning of each new architecture is eliminated.
(iv) \textbf{Continual learning}: The NCWNO can continually learn the solution operators of multiple physical systems without catastrophically forgetting the previous tasks. The NCWNO also provides a positive transfer of the previously learned physics on new parametric PDEs, allowing the learning of new parametric PDEs from reduced data and increased efficiency. 

The major contributions of the proposed method to the existing literature include: (1) It is the first foundational operator learning model proposed for simulations and applications in scientific computing that can learn solution operators of different parametric PDEs using a single combinatorial model (i.e., maps between a set of infinite-dimensional function spaces), as opposed to the neural operators (which learn the solution operator of only a single parametric PDE) or the neural networks (which learn a functional map between two finite-dimensional vector space). It also continually learns the solution operators of parametric PDEs without performance degradation on the old tasks. (2) With the help of the pre-trained foundation model, an efficiency of $>$50\% is observed on both the data requirement and computational time.
We consider a number of numerical examples representing various phenomena in various domains of computational physics to test the performance of the proposed NCWNO framework. Together with the advances in graphics-based computing power and the ability to learn multitask problems, the proposed NCWNO will be able to aid the scientific community in modeling and simulation of complex nonlinear processes under multiple operating conditions across engineering design and control, computational physics, and computational biology.

\section{Results}
\begin{figure}[t!]
    \centering
    \includegraphics[width=\textwidth]{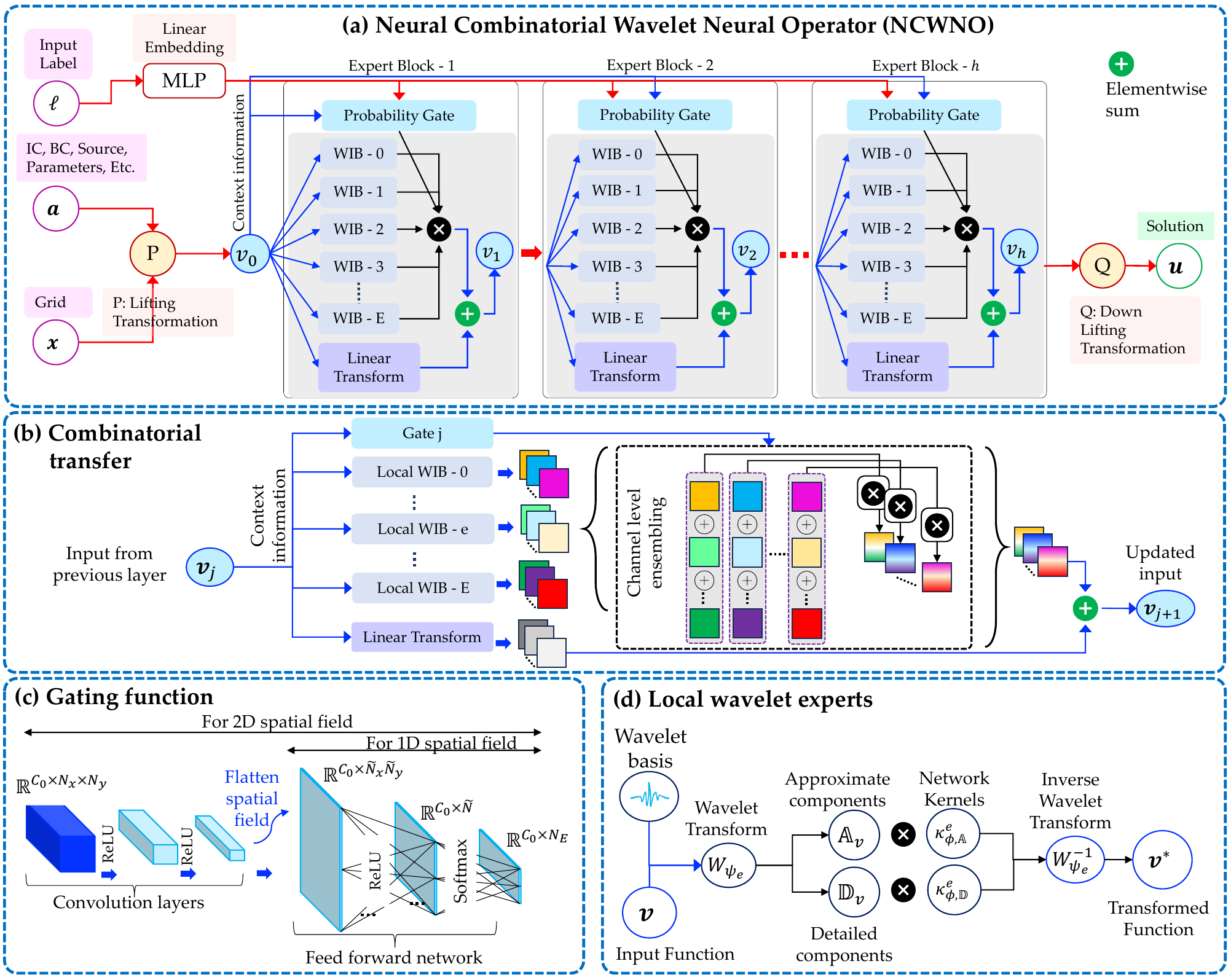}
    \caption{\textbf{Schematic architecture of the neural combinatorial wavelet neural operator (NCWNO)}. (\textbf{a}) The overview of the network architecture. The context function is derived from the input parameters and iteratively updated through expert wavelet integral blocks, where the outputs from local experts are ensembled using model probabilities from gating functions. The network first simultaneously learns a set of parametric PDEs. During combinatorial transfer learning of new parametric PDEs, the kernel parameters of gating functions are updated while fixing the kernels of expert integral blocks. (\textbf{b}) Based on the ensembling probabilities from gating functions, a local channel-level averaging is performed between the expert outputs, indicated by the color gradients. (\textbf{c}) A gating function with softmax operation at the output is applied to the input function, yielding the channel-level ensembling probabilities of each local expert. (\textbf{d}) The input is decomposed using undecimated DWT, followed by convolution with the network kernels and inverse wavelet transform on the convolved input.}
    \label{fig_methodology}
\end{figure}

\subsection{The overview of NCWNO}
Figure \ref{fig_methodology} shows an overview of the proposed NCWNO framework. The deployment of NCWNO comprises two main stages. The first stage involves training the NCWNO model on multiple physical systems for different input parameters, yielding the initial foundation model or the universal solution operator of multiple parametric PDEs. The second stage involves combinatorial transfer learning on new tasks that involve finetuning only the gated network instead of retraining the expert layers on relatively smaller datasets of new parametric PDEs. 
As illustrated in Fig. \ref{fig_methodology}(a), the NCWNO takes input parameters $\bm{a}$, the grid $\bm{x}$ on which the solutions are required, and the corresponding PDE label $\ell$. The input parameters $\bm{a}$ and the grid $\bm{x}$ are uplifted to a higher dimension using a transformation P. The expert integral blocks take the lifted context function $\bm{v}_0$, whereas the gating functions take both the lifted context function and the corresponding PDE label. Each expert integral block is accompanied by a gating function. The expert integral blocks consist of an ensemble of local wavelet experts whose task-specific contributions in predicting the PDE solutions are decided by the output probabilities from the gating functions. The final output obtained from $h$ sequential expert integral blocks is transformed to the prediction using a local transformation Q. 
As demonstrated in Fig. \ref{fig_methodology}(b), given the outputs from local wavelet blocks and the channel-level probabilities from the gating function, a channel-level ensembling is performed, indicated by the color gradients. Since the ensembling probabilities vary across channels, the NCWNO can consider the interactions between channels more effectively than simple model averaging. 
Figure \ref{fig_methodology}(c) shows the gating network, which is designed as a sequential network of convolution and feed-forward neural networks. Figure \ref{fig_methodology}(d) shows the methodology of the local wavelet experts, where the spectral decomposition of the input context function is performed using wavelet transform followed by kernel convolution in the wavelet domain. The convolved input is finally transformed back using inverse wavelet transform.

\subsection{Learning foundation model on multiple parametric PDEs}
The NCWNO first learns a foundation model by approximating the solution operators of multiple parametric PDEs concomitantly. For forthcoming PDEs, NCWNO performs a combinatorial transfer of knowledge from the foundation model to new tasks. 
In this section, we evaluate the performance of the NCWNO by learning the foundation model on five time-dependent parametric PDEs. The dataset for each parametric PDE contains 1000 paired functional input-output. The input contains different initial conditions, and the output contains the functional solution maps of the initial conditions. Both 1D and 2D time-dependent PDE problems are considered, where the aim is to learn a single foundation model that can accurately predict the time evolution of the solutions of multiple parametric PDEs. 
The set of five 1D time-dependent PDEs contains the 1D Allen Cahn equation, 1D Nagumo equation, 1D Burgers equation, 1D Heat equation, and 1D Wave equation. The set of five 2D time-dependent PDEs contains the 2D Advection equation, 2D Nagumo equation, 2D Allen Cahn equation, 2D Heat equation, and 2D Navier-Stokes equation. The details of the synthetic data generation are provided in the methods section.   
\begin{figure}[t!]
    \centering
    \includegraphics[width=\textwidth]{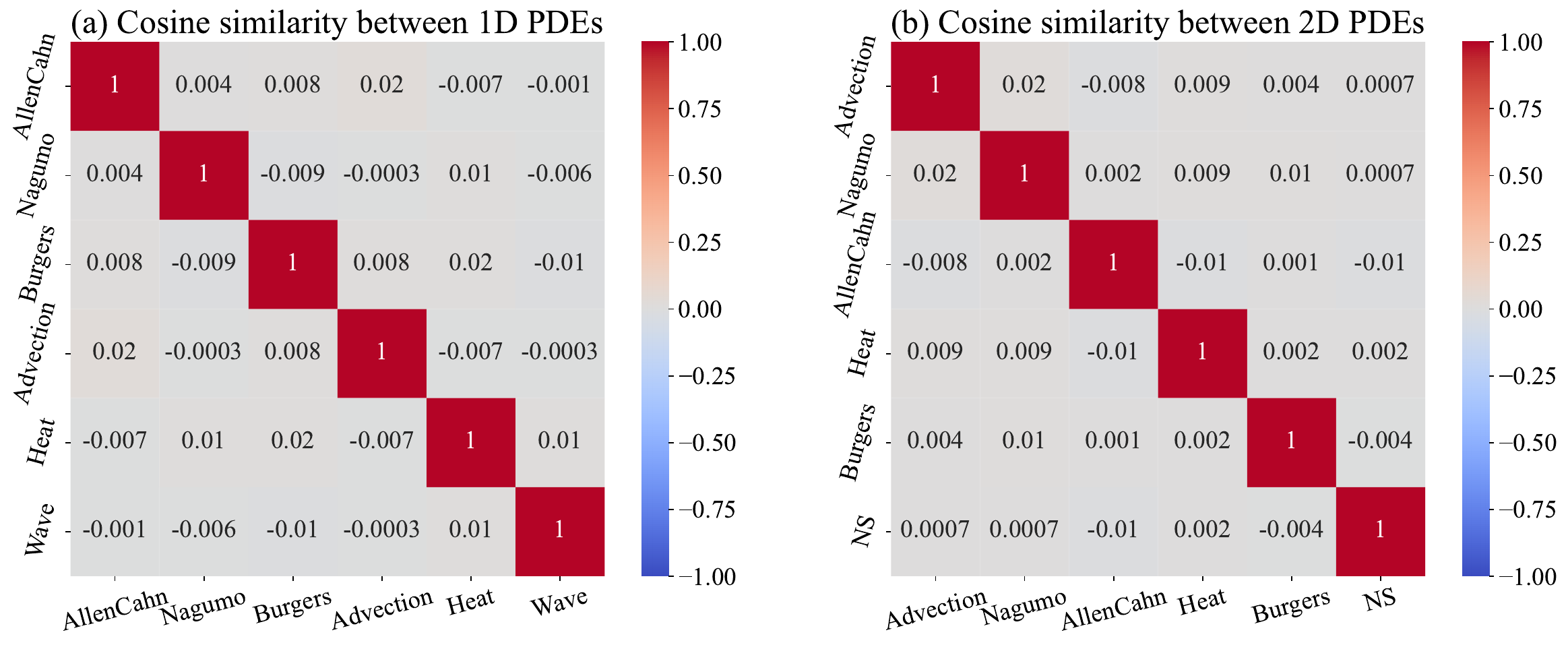}
    \caption{\textbf{Cosine similarity between the datasets of different parametric PDEs}. The dataset contains the time-evolving solution map of parametric PDEs for different initial conditions simulated as a Gaussian random field. A value of 1 indicates a perfect match between tasks, and a value of 0 indicates a completely different task.}
    \label{fig:similarity}
\end{figure}
To measure the dissimilarity between the parametric PDEs, we measure the pairwise cosine similarity $\mathcal{S}(\mathcal{X},\mathcal{Y})$ between the datasets $\mathcal{X}$ and $\mathcal{Y}$ using the formula: $\mathcal{S}(\mathcal{X},\mathcal{Y}) = (\mathcal{X}^T \mathcal{Y}) / (\| \mathcal{X} \| \| \mathcal{Y} \|)$, where $\|\cdot\|$ denotes the Frobenius Norm. It is evident from the $\mathcal{S}(\mathcal{X},\mathcal{Y})$ values in Fig. \ref{fig:similarity} that the tasks are significantly different from each other.
In the set of 1D parametric PDEs, the aim is to learn the universal solution operator $\mathcal{D}^{\mathcal{N}_{1:5}}_{\bm{\theta}}: u\vert^{1:5}_{\Omega \times [0,10]} \mapsto u\vert^{1:5}_{\Omega \times [11,40]}$, that maps the solutions of the 1D parametric PDEs at first ten time-steps to the solutions at next thirty time steps for arbitrary initial conditions. For 2D time-dependent parametric PDEs, we learn the universal solution operator $\mathcal{D}^{\mathcal{N}_{1:5}}_{\bm{\theta}}: u \vert^{1:5}_{\Omega \times [0,10]} \mapsto u \vert^{1:5}_{\Omega \times [11,20]}$, that maps the solutions of the 2D parametric PDEs at first ten-time steps to the solutions at next ten time steps. 
\begin{figure}[ht!]
    \centering
    \includegraphics[width=\textwidth]{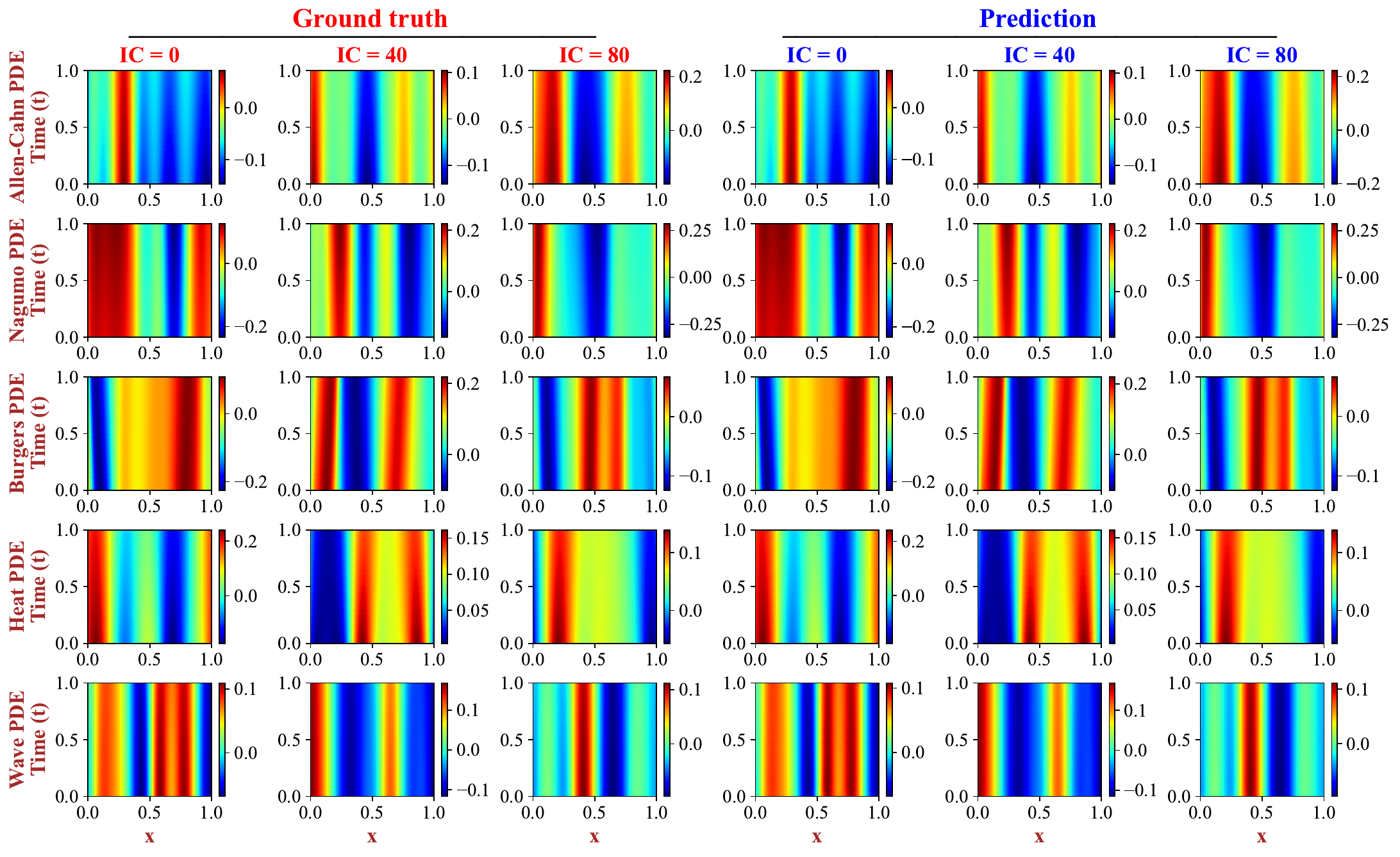}
    \caption{\textbf{Learning foundation model for the 1D time-dependent parametric PDEs}. A single NCWNO is trained concomitantly on the Allen-Cahn, Nagumo, Burgers, Heat, and Wave equations. The NCWNO learns the universal solution operator of the underlying PDEs. Ground truth versus the predictions of the trained NCWNO model for three representative initial conditions are illustrated. The solutions are obtained in a time-marching fashion.}
    \label{fig:combined_1d}
\end{figure}

\begin{figure}[ht!]
    \centering
    \includegraphics[width=\textwidth]{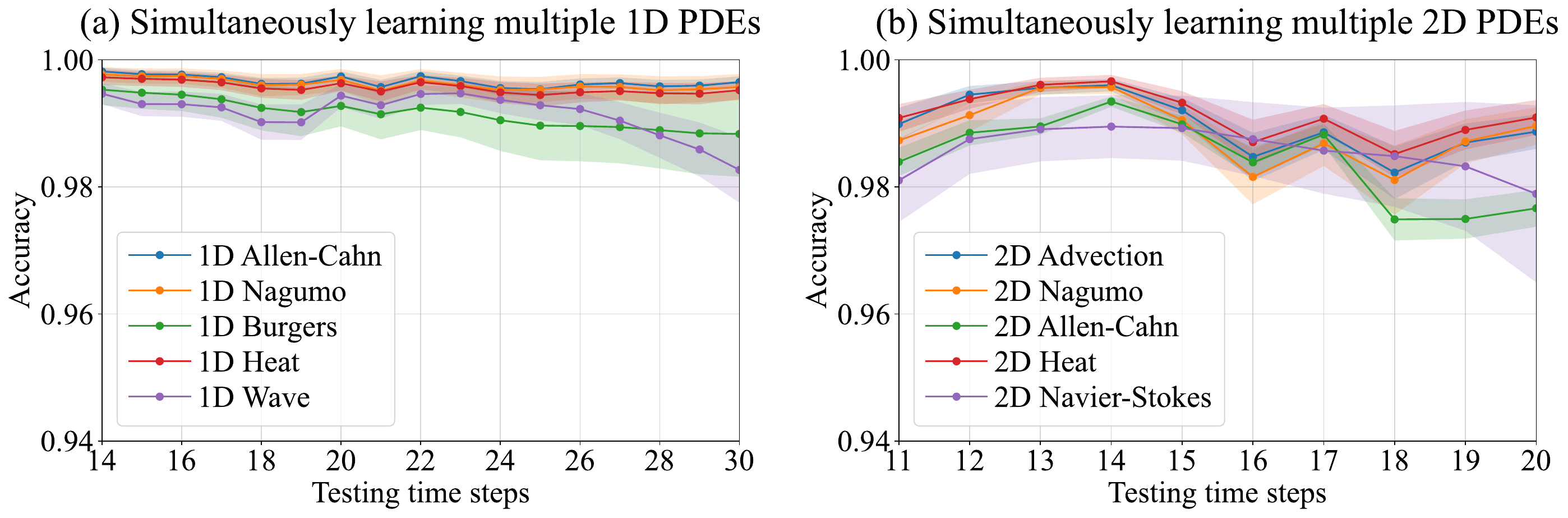}
    \caption{\textbf{Predictive accuracy of temporal predictions of each parametric PDE after learning all the tasks in a single go}. The testing sample size of each parametric PDE contains 100 different initial conditions. The shaded region indicates the 95\% confidence interval estimated over the entire sample size.  The accuracy is measured by subtracting the relative $L^2$ error from 1.}
    \label{fig:accuracy}
\end{figure}

The time marching solutions of the underlying 1D time-dependent parametric PDEs predicted using the trained NCWNO for three different representative initial conditions are shown in Fig. \ref{fig:combined_1d}. For each PDE, the y-axis shows the time evolution, and the x-axis shows the spatial grid. The PDE labels are marked along the y-axis. In Fig. \ref{fig:combined_1d}, it can be observed that the NCWNO predicted results of each PDE for all three different initial conditions almost exactly emulate the ground truths. We further estimate the accuracy of the temporal predictions against ground truth as follows: $\operatorname{acc}:= 1 - (\|\bm{u} - \bm{u}^{*}\|/\|\bm{u}\|)$, where $\bm{u}$ and $\bm{u}^{*}$ denote the ground truth and predictions from trained NCWNO model. The higher the value of the accuracy metric, the better the predictions. The statistics of the accuracy of the NCWNO predictions against the ground truth across different time steps for 100 test samples are provided in Fig. \ref{fig:accuracy}(a), where the shaded region indicates the 95\% confidence bound. 
The 95\% confidence bound is obtained by estimating the standard deviation of the accuracy across the sample size. Figure \ref{fig:accuracy}(a) indicates that the accuracy metric for the NCWNO predictions of all 1D parametric PDEs is close to 1. In particular, we obtain an average predictive accuracy of $0.9965 \pm 0.0015$ for Allen-Cahn equation, $0.9961 \pm 0.0023$ for Nagumo equation, $0.9916 \pm 0.0051$ for Burgers equation, $0.9955 \pm 0.0021$ for Heat equation, and $0.9911 \pm 0.0053$ for Wave equation. 

\begin{figure}[t!]
    \centering
    \includegraphics[width=\textwidth]{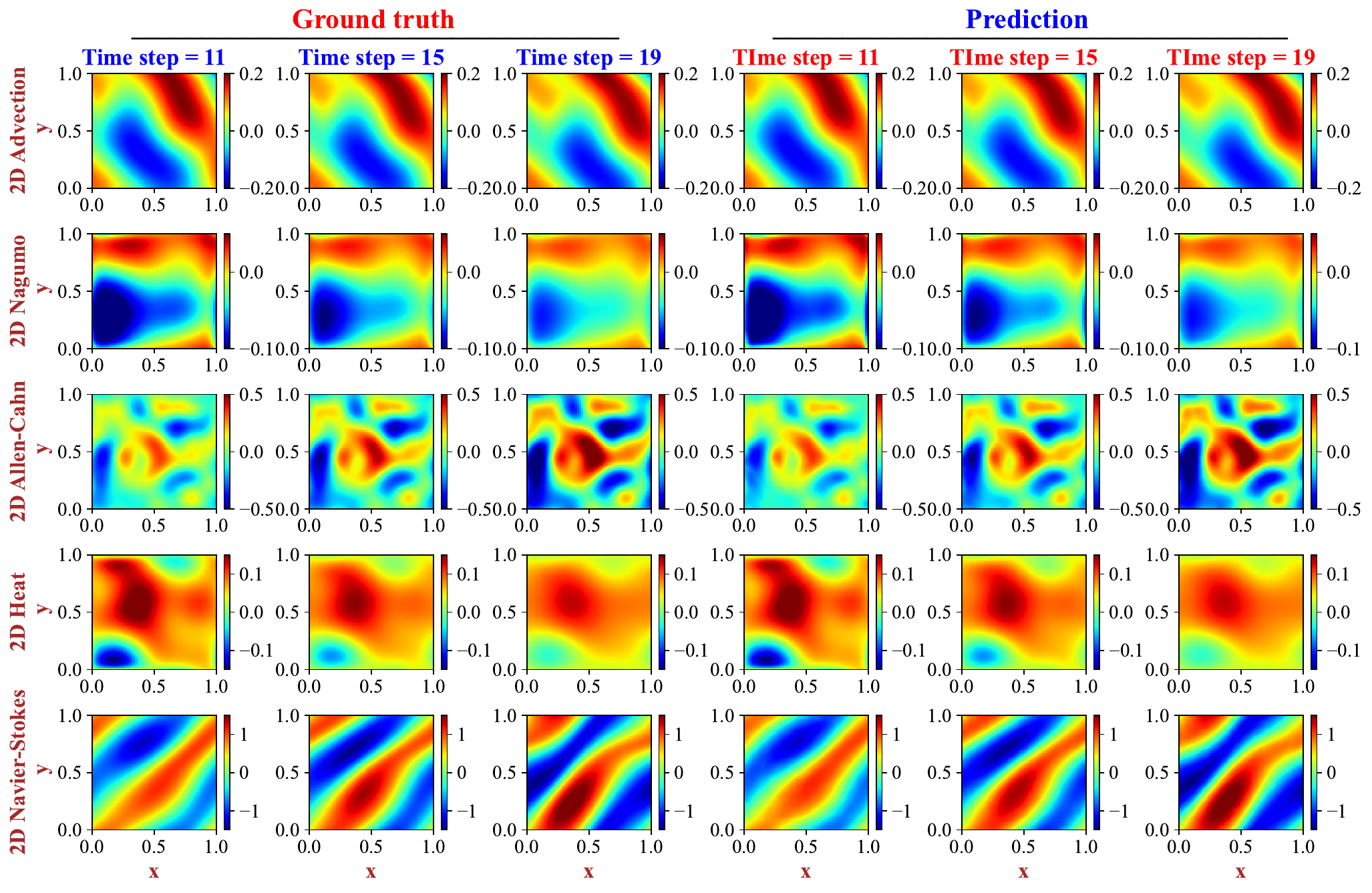}
    \caption{\textbf{Learning foundation model for the 2D time-dependent parametric PDEs}. The foundation model learns the universal solution operator of the Advection, Nagumo, Allen-Cahn, Heat, and Navier-Stokes equations. The time evolution of the predictions of the trained foundation model against the ground truth for a given initial condition is portrayed.}
    \label{fig:combined_2d}
\end{figure}
Following a similar procedure to the 1D PDEs, we learn a foundation model for the 2D parametric PDEs concomitantly. The time-marching solutions of underlying 2D time-dependent PDEs predicted using the trained NCWNO model for a given initial condition at three different time steps are illustrated in Fig. \ref{fig:combined_2d}. The x and y-axis denote the 2D spatial grid of the solution space. The time steps are mentioned in the subplot titles. We observe that the prediction results almost exactly reproduce the ground truth results, indicating the robustness of the learned NCWNO model. The statistics of the predictive accuracy of the time-marching solutions averaged over 100 different initial conditions are illustrated in Fig. \ref{fig:accuracy}(b). In terms of the accuracy metric, we observe an average predictive accuracy of $0.9884 \pm 0.0071$ on Advection equation, $0.9867 \pm 0.0087$ on Nagumo equation, $0.9814 \pm 0.0101$ on Allen-Cahn equation, $0.9899 \pm 0.0062$ on Heat equation, and $0.9856 \pm 0.0084$ on Navier-Stokes equation. 
The ability to learn the physics and solution operators of multiple heterogeneous PDEs and predict time-marching solutions for different input parameters is truly a novelty of the proposed NCWNO. 
We also compare the performance of the NCWNO against existing robust task-specific operator learning frameworks like DeepONet, FNO, and WNO, which can learn the solution operator of only a given PDE. For a detailed comparison, see Appendix \ref{sec:benchmark}. We consider nonlinear and chaotic problems from the literature and show that the proposed NCWNO outperforms the existing DeepONet, FNO, and WNO in all the respective examples. 
In the next section, we explore the continual learning performance of the proposed NCWNO on different parametric PDEs.

\begin{figure}[b!]
    \centering
    \includegraphics[width=\textwidth]{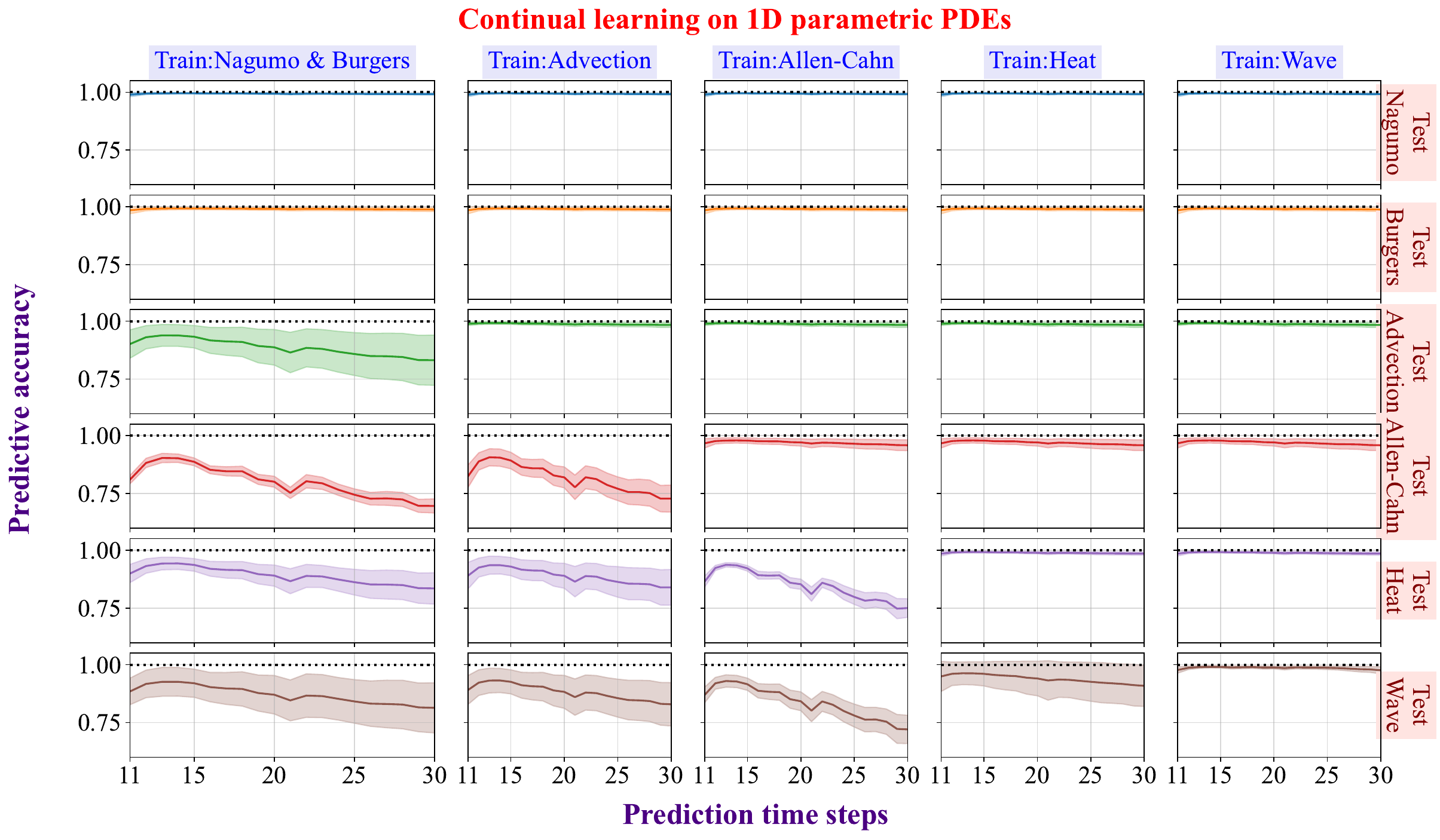}
    \caption{\textbf{Combinatorial transfer learning of solution operators of 1D parametric PDEs}. The foundation model is trained on Nagumo and Burgers PDE, later used to sequentially train on up to 4 new parametric PDEs (indicated by columns), and sequentially test on each task (indicated by rows). For example, the third column indicates the foundation model's performance on all tasks after being sequentially trained on the Advection and Allen-Cahn equations. Similarly, the right column indicates the foundation model's performance on all tasks after being sequentially trained on parametric PDEs 3 to 6.}
    \label{fig:acc_1d}
\end{figure}
\subsection{Performance on continual learning of parametric PDEs}
\begin{figure}[t!]
    \centering
    \includegraphics[width=\textwidth]{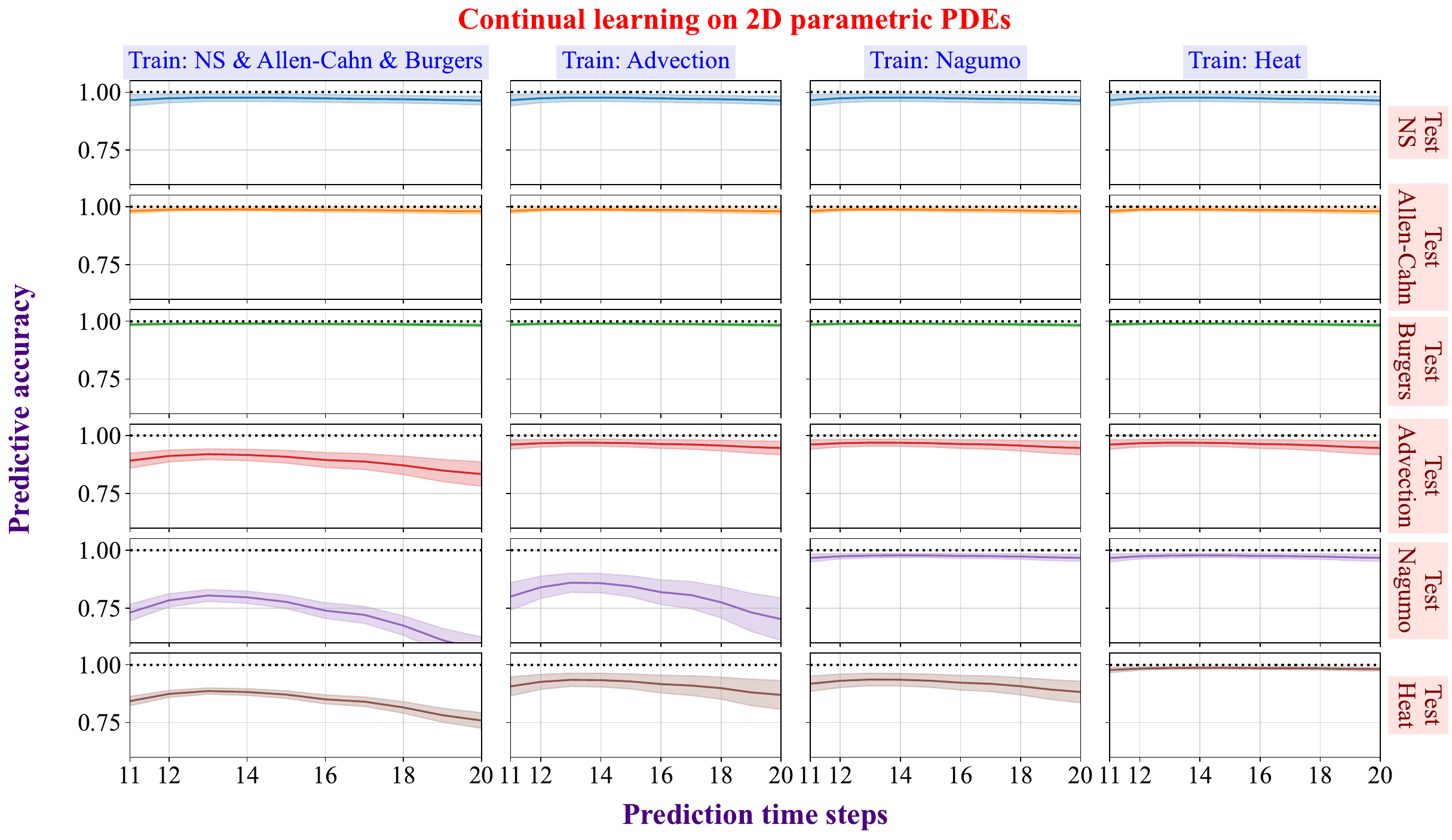}
    \caption{\textbf{Combinatorial transfer learning of solution operators of 2D parametric PDEs}. The foundation model is trained on Navier-Stokes, Allen-Chan, and Burgers PDE, later used to sequentially train on up to 3 new parametric PDEs (indicated by columns), and sequentially test on each task (indicated by rows). For example, the rightmost column indicates the foundation model's performance on all tasks after being trained sequentially on Advection, Nagumo, and Heat PDE. }
    \label{fig:acc_2d}
\end{figure}

Here, we intend to continually learn the solution operators of the parametric PDEs for predicting the time evolution of the solutions of new unseen physical systems. We evaluate the robustness of the proposed NCWNO against catastrophic forgetting by sequentially learning up to four different parametric PDEs. We draw the readers' attention to the fact that continually learning solution operators of time-dependent parametric PDEs is more challenging than the classification and regression of stationary PDEs. 
We first demonstrate the continual learning performance of the NCWNO on 1D parametric PDEs. We instantiate a foundation model for the 1D parametric PDEs by learning the universal solution operator of the 1D Nagumo and 1D Burgers equation for different initial conditions. We denote the underlying operator as $\mathcal{D}^{\mathcal{N}_{1:2}}_{\bm{\theta}}: u \vert^{1:2}_{\Omega \times [0,10]} \mapsto u \vert^{1:2}_{\Omega \times [11,40]}$, that maps the solutions of 1D Nagumo and Burgers equation at first ten-time steps in the domain $\Omega \in \mathbb{R}^d$ to the solutions at next thirty time steps. 
With the pre-trained foundation model, we perform combinatorial transfer learning (sequentially) on unseen 1D Advection, 1D Allen-Cahn, 1D Heat, and 1D Wave PDE. During combinatorial transfer learning of new parametric PDEs, we adapt the foundation model by fine-tuning only the parameters of gating functions. 
For a new parametric PDE $\mathcal{N}_{\tau}$, we learn the solution operator $\mathcal{D}^{\mathcal{N}_{1:2} \cup \mathcal{N}_\tau}_{\bm{\theta}}: u \vert^{\{1:2\} \cup \{\tau \}}_{\Omega \times [0,10]} \mapsto u \vert^{\{1:2\} \cup \{\tau \}}_{\Omega \times [11,40]}$, which predicts the solutions of the new PDE $\mathcal{N}_{\tau}$ while retaining the knowledge gained previously. 
We use the previously defined accuracy metric as a measure of accuracy in continual learning. 
The predictive accuracy of the proposed NCWNO in continuously adapting to new systems is illustrated in Figs. \ref{fig:acc_1d}. The shaded region denotes 95\% confidence levels averaged over 100 different initial conditions. We observe that the proposed NCWNO learns new solution operators without forgetting the previously learned operators (as indicated by the upper triangular figure matrix). Interestingly, even for tasks that are quite different from each other (see Fig. \ref{fig:similarity} for similarity index), the NCWNO performs a positive transfer of the foundation model's knowledge to new parametric PDEs without sacrificing accuracy on the previously learned PDEs. For the 2D examples shown in Fig. \ref{fig:acc_2d}, the foundation model is initiated by training it on 2D Navier-Stokes, Allen-Chan, and Burgers equations. Subsequently, it is sequentially fine-tuned on the Advection, Nagumo, and Heat equations. In this case also, the proposed NCWNO learns the solution operator of new systems while retaining knowledge of the previously learned systems. Overall, the ability to sequentially learn solution operators of new PDEs while retaining the knowledge about the previously learned solution operators clearly indicates the proposed NCWNO's robustness against catastrophic forgetting; this is even true for systems that are dissimilar.
This indicates that the channel-level ensembling of local wavelet experts helps the NCWNO to continually adapt and learn the solution operators of new systems.

\subsection{Reduced training cost on new parametric PDEs} 
The ability to generalize to new downstream tasks with increased resource efficiency is the main aim of using foundation models. In Table \ref{tab:epochs}, it is evident that with a pre-trained foundation model, the NCWNO reduces the number of training epochs for new tasks by more than 3 times and the labeled data requirements and computational time by a factor of 2. The ability to perform combinatorial transfer learning of new dissimilar parametric PDEs (evident from Fig. \ref{fig:similarity}) with highly accurate prediction (evident in Fig. \ref{fig:acc_1d}) and reduced memory resources (indicated in Table \ref{tab:epochs}) is one of the key advantages of NCWNO. 
\begin{table}[ht!]
    \centering
    \caption{Training costs for training initial foundation model and combinatorial transfer learning of new time-dependent parametric PDEs. The foundation model is trained on Task 1 \& 2 simultaneously.}
    \label{tab:epochs}
    \begin{tabular}{llllll}
    \hline
    & Task 1 \& 2 & Task 3 & Task 4 & Task 5 & Task 6 \\
    \hline
    1D Examples & Nagumo \& Burgers & Advection & Allen-Cahn & Heat & Wave \\
    \hline
    Epochs & 150 & 50 & 50 & 50 & 50 \\
    Time (in sec.) & 196.6744 & 105.5749 & 105.5777 & 105.6934 & 105.1112 \\
    Data pair & 1000 & 500 & 500 & 500 & 500 \\
    \hline
    2D Examples & Advection \& Nagumo & Allen-Cahn & Heat & Burgers & Navier-Stokes \\
    \hline
    Epochs & 100 & 50 & 50 & 50 & 50 \\
    Time (in sec.) & 678.0701 & 264.2409 & 264.5483 & 265.9591 & 264.2733 \\
    Data pair & 1000 & 500 & 500 & 500 & 500 \\
    \hline
    \end{tabular}
\end{table}

\subsection{Effect of the size of local wavelet experts}
Table \ref{tab:ablation} provides an overview of the effect of the number of local wavelet experts on the robustness of the NCWNO against catastrophic forgetting of old tasks. The values in the Table \ref{tab:ablation} represent the relative $L^2$ norm of the prediction error estimated as $\epsilon = \|\bm{u} - \bm{u}^{*}\| / \|\bm{u}\|$, with $\bm{u}^{*}$ and $\bm{u}$ being the predictions and ground truth, respectively. It can be seen that increasing experts increases the performance of the NCWNO on both the foundation tasks (Tasks 1 \& 2) and continual tasks (Tasks 3-6). 
Although the expert population plays a constructive role in making the NCWNO a successful continual learner, the computation time also increases accordingly. Here, 4, 7, 10, 13, and 16 experts take approximately 57s, 75s, 105s, 173s, and 210s of compute time per epoch. 
\begin{table}[ht!]
    \centering
    \caption{Effect of the number of experts on catastrophic forgetting of old tasks. This case study is exemplified by the 1D parametric PDEs. The errors denote the mean relative $L^2$ error average over 100 different initial conditions.}
    \label{tab:ablation}
    \begin{tabular}{lllllll}
    \hline
    \multirow{2}{*}{Expert size} & \multicolumn{2}{l}{Foundation learning on Task 1 \& 2} & \multicolumn{4}{l}{Sequential learning on Task 3, 4, 5, \& 6} \\
    \cmidrule(r){2-3} \cmidrule(r){4-7}
    & Nagumo & Burgers & Advection & AllenCahn & Heat & Wave \\
    \hline
    4 & 0.0359 & 0.0441 & 0.3077 & 0.4993 & 0.2784 & 0.6041 \\
    7 & 0.0050 & 0.0100 & 0.1244 & 0.0346 & 0.0208 & 0.0149 \\
    10 & 0.0066 & 0.0098 & 0.0120 & 0.0301 & 0.0113 & 0.0140 \\
    13 & 0.0036 & 0.0088 & 0.0121 & 0.0282 & 0.0083 & 0.0127 \\
    16 & 0.0036 & 0.0059 & 0.0120 & 0.0149 & 0.0081 & 0.0126 \\
    \hline
    \end{tabular}
\end{table}

\section{Discussion}
In this work, we propose the NCWNO as a foundation model for the scientific computing of parametric PDEs. The NCWNO can learn the nonlinear solution operators of multiple parametric PDEs from a combined dataset of multiple physical systems. This is facilitated by the modular learning of local features of the global PDE kernel through a set of local wavelet experts. With the pre-trained foundation model, the NCWNO can continually approximate the nonlinear solution operators of new parametric PDEs by local channel-level ensembling of the previously learned features across pre-trained parametric PDEs. 
The ability to predict solutions of multiple heterogeneous nonlinear PDEs for multiple parametric inputs without catastrophic forgetting of the dynamics of old parametric PDEs is a novel contribution of the proposed NCWNO to the existing literature.
Compared to existing neural operator learning techniques (which can learn the operator of only a single parametric PDE), the proposed NCWNO can learn the universal solution operator for a set of parametric PDEs, further reducing the need for repeated training of neural operators for different physical systems. This is another novel contribution of this work to the existing literature.
Through the composition of a finite number of local wavelet experts and a local channel-level ensembling mechanism between the local expert blocks, NCWNO encapsulates the PDE-specific operator learning, multi-physics learning, and continual learning in a single framework. 

We numerically illustrated the robustness of the proposed NCWNO for multiple physics learning and continual learning without catastrophic forgetting on six different parametric PDEs. The dataset for each parametric PDE contains 1000 different input-output combinations. The results indicate that the NCWNO can predict the dynamics of multiple complex nonlinear processes under arbitrary initial conditions. The results also demonstrate that continual learning of new parametric PDEs using the knowledge of previously learned PDEs increases the accuracy of the predictions on new PDEs while reducing the memory and computational times by a factor of 2. This clearly illustrates the importance of combinatorial transfer in improving the accuracy of the predictions and generalizing to unseen input parameters with minimal fine-tuning. 
We also compare the robustness of NCWNO against existing task-specific neural operators (see Appendix \ref{sec:benchmark}). The comparative results show that the channel-level ensembling of local wavelet experts in the NCWNO outperforms the existing neural operators in a task-specific platform without overfitting the unseen data. 

Overall, this work explores two key aspects of scientific machine learning. The first aspect is associated with the fidelity of machine learning models in multiple operating environments, where we ask whether a single machine learning model can be deployed for performing multiple tasks without fine-tuning during inference. The second aspect asks whether the deployed model can acquire, assimilate, and retain new information, as a human brain does without forgetting the old tasks. We observe that the NCWNO satisfactorily achieves both objectives. Hence, we envisage that the NCWNO can provide a new direction for AI-driven automation of scientific and engineering systems under evolving operating conditions.

Having demonstrated the performance of the NCWNO for learning the universal solution operator of multiple parametric PDEs and adapting to new PDEs without catastrophic forgetting, there are certain limitations of the proposed NCWNO.
Although the kernel convolutions in the local wavelet experts preserve the properties of zero-shot super-resolution, i.e., train on a different resolution and predict on another resolution, due to the feedforward gating function, the NCWNO in its current can not perform zero-shot super-resolution prediction. Although instantiating the gating function as a wavelet operator, tailored to predict the mixing probabilities of the local experts, could be a possible solution, how to devise an appropriate wavelet operator algorithm for estimating model probabilities is yet not clear. 
The NCWNO also assumes that the task identities/PDE labels are known a priori. A practical future endeavor will involve introducing a reflexive mechanism in the NCWNO that can track the reflexes of experts and discover the task identities. This will relax the need for PDE labels in the gating mechanism, allowing the deployment of NCWNO to problems with unknown task identities. 

Possible extensions of the NCWNO include the development of multiscale NCWNO and physics-informed NCWNO. For a problem with a very large domain, performing a full microscale simulation often becomes computationally uneconomic. In such a case, we can accelerate the modeling of multiscale parametric PDEs by learning the multiscale operators using NCWNO. 
Further, most of the natural processes are multiphysics in nature. While simulating an individual aspect of a physical process is relatively easier, the simulation of the interactions between different physics is extremely challenging. However, by biasing the local wavelet experts to satisfy the physics of an individual aspect of a multi-physics process, we can derive the physics-informed NCWNO for efficiently performing multiphysics simulation. These will have direct applications in various horizons of climate modeling, fluid-structure interaction, particle physics, thermodynamics, and material science.

\section{Methods}
The proposed NCWNO generalizes the existing concepts of the neural operator to simultaneous and continual learning of multiple heterogeneous physical systems by seamless integration of the strengths of modular learning and transfer learning with the benefits of kernel parameterization in the wavelet domain. The NCWNO introduces the local wavelet experts for kernel parameterization using different local wavelet basis functions. Each local expert learns a distinct feature that is commonly shared by multiple heterogeneous parametric PDEs. The NCWNO utilizes a gating mechanism for modular learning by locally ensembling the channels of each local wavelet expert. By leveraging the previously learned building blocks of each local wavelet expert, the proposed NCWNO learns to predict new parametric PDEs. The mathematical descriptions of the proposed framework are discussed below. 

\subsection{Neural Combinatorial Wavelet Neural Operator (NCWNO)}
The schematic details of the proposed NCWNO is provided in Fig. \ref{fig_methodology}. For a comprehensive description, we consider the domain of the parametric PDEs to be $\Omega$ with the appropriate boundary $\partial \Omega$. In the domain $\Omega$, we denote the parameter and solution spaces of a PDE as $\mathcal{A}:= \mathcal{C}(\Omega;\mathbb{R}^{d_a})$ and $\mathcal{U}:= \mathcal{C}(\Omega;\mathbb{R}^{d_u})$ of functions taking values in $\mathbb{R}^{d_a}$ and $\mathbb{R}^{d_u}$, respectively ($\mathcal{C}$ is continuous space). Further, we denote $x \in \Omega$ to be a spatial point in $\Omega$ and $t$ to be a time coordinate. 
We emphasize that the parameter space $\mathcal{A}$ represents a collective space of the source term $f(x, t): \Omega \times \mathcal{C} \mapsto \mathcal{C}$, the initial condition $u(x,t_0): \Omega \times \mathcal{C} \mapsto \mathcal{C}$, the boundary condition $u(\partial \Omega, t): \partial \Omega \times \mathcal{C} \mapsto \mathcal{C}$, and the PDE coefficients $\lambda(x): \Omega \mapsto \mathcal{R}$, opposed to only the PDE coefficients. On the contrary, $\mathcal{U}$ contains the usual solutions $u(x, t): \Omega \times \mathcal{C} \mapsto \mathcal{C}$ of the underlying parametric PDE. 
We assume that there exists a differential operator $\mathcal{N}:\mathcal{A}\times \mathcal{U} \mapsto 0$. Then assuming that a unique solution $\bm{u} \in \mathcal{U}$ exists for every $\bm{a} \in \mathcal{A}$, the NCWNO seek to learn the solution/anti-derivative operator $\mathcal{D}: \bm{a} \mapsto \bm{u}$, by parameterizing the integral kernel of the operator $\mathcal{D}$ with a finite-dimensional parameter space $\bm{\theta}$ as, 
\begin{equation}\label{eq:anti_derivative}
    \mathcal{D} : \mathcal{A} \times {\bm{\theta}} \mapsto \mathcal{U} .
\end{equation}
We assume that the data are recorded on a $d$ point discretized domain $\Omega^d = \{x_1, \ldots, x_d\}$. Assuming that we have a set of size $N$ of the input-solution pair on the discretized domain, represented as $\{ \mathbf{a}_{j} \in \mathbb{R}^{d \times d_{a}}, \mathbf{u}_{j} \in \mathbb{R}^{d \times d_{u}} \}_{j=1}^N$, the complete process of learning the mapping $\bm{u}(x) = \mathcal{D}(\bm{a})(x)$ for an input $\bm{a}(x)$ can be represented as, 
\begin{equation}\label{eq:operator_map}
    \bm{u}(x)= {\rm{Q}} \left( q_h\left( \ldots q_j\left( \ldots q_0\left( {\rm{P}}(\bm{a}(x)) \right) \ldots \right) \ldots \right) \right), \;\; x \in \Omega^d , 
\end{equation}
where ${\rm{P}}: \mathbb{R}^{d_a} \mapsto \mathbb{R}^{d_v}$ ($\mathbb{R}^{d_a} < \mathbb{R}^{d_v}$) is a transformation that encodes the input parameters $\bm{a}$ and the gird information $x$ to the context information $\bm{v}$ and ${\rm{Q}}: \mathbb{R}^{d_v} \mapsto \mathbb{R}^{d_u}$ is a transformation that maps the transformed context information $\bm{v}_h$ to the solution $\bm{u}$. 
Denoting $\bm{v}_0(x) = {\rm{P}}(\bm{a}(x))$, the hidden iterations $q_j: \mathbb{R}^{d_v} \mapsto \mathbb{R}^{d_v}$ are expressed as \cite{li2020fourier},
\begin{equation}\label{eq:iteration}
    \bm{v}_{j+1}(x) = \Gamma \left( \left( \mathcal{K}(\bm{a}; \phi_j \in \bm{\theta}) \bm{v}_{j}\right)(x) + (g \bm{v}_{j})(x) \right), \;\; x \in \Omega^d , \;\; j \in [1,h],
\end{equation}
where $\Gamma(\cdot): \mathbb{R} \to \mathbb{R}$ represents the non-linear activation function, $g: \mathbb{R}^{d_{v}} \to \mathbb{R}^{d_{v}}$ represents a linear transformation, $\mathcal{K}: \mathcal{A} \times {\bm{\theta}} \to \mathcal{U}$ is the integral operator on $\mathcal{C} (\Omega ; \mathbb{R}^{d_{v}})$, and $\phi_j \in \bm{\theta}$ represents a subset of network parameters. Motivated by the nonlinear Hammerstein-type integral equation: $u(x) = \int_{\Omega} k(x,\xi) f(u(x), \xi) d\xi$, from classical functional analysis \cite{yosida2012functional}, the kernel integral operator $\mathcal{K}(\bm{a}(x); \phi_j)$ is defined as,
\begin{equation}\label{eq:integral}
    \left(K(\bm{a}; \phi_j \in \bm{\theta}) \bm{v}_{j}\right)(x) = \int_{\mathbb{R}^{d}} k \left(\bm{a}(x), \bm{a}(\xi), x, \xi; \phi_j \in \bm{\theta} \right) \bm{v}_{j}(\xi) d \xi; \;\; x \in \Omega^d, \;\; j \in [1,h],
\end{equation}
where $k(\cdot; \phi_j)$ is called the kernel of the above integral operator. In the simultaneous learning strategy of multiple heterogeneous parametric PDEs, the kernel $k \in \mathcal{C}(\Omega; \mathbb{R}^{d_v})$ learns the features that are broadly useful across diverse physical systems, represented a set of parametric PDEs. We aim to learn those shared features by approximating the true kernel function $k \in \mathcal{C}(\Omega; \mathbb{R}^{d_v})$ through local ensembling between $d_e$ number of local wavelet integral blocks.
We denote the kernels of local wavelet integral blocks as $k^{e}_{\phi_j}:=k(\cdot; \phi_{j}^{e})$, which are averaged in the channel-level using the model probabilities $\bm{\beta} \in \mathbb{R}^{d_e}$. The Eq. \eqref{eq:integral} is redefined as, 
\begin{equation}\label{eq:moe_integral}
    \left(\mathcal{K}(\bm{a}; \phi_j \in \bm{\theta}) \bm{v}_{j}\right)(x) = \sum_{e=1}^{d_e} \left[ \bm{\beta}^{e}_{j} \cdot \int_{\mathbb{R}^{d}} k^{e}_{\phi_j} \left(\bm{a}(x), \bm{a}(\xi), x, \xi \right) \bm{v}_{j}(\xi) d \xi \right]; \;\; x \in \Omega^d, \;\; j \in [1,h],
\end{equation}
where $\bm{\beta}^{e}_{j} = \sigma(p_{\beta}(e \mid \bm{v}_j(x) \ell; \bm{\theta}_{\beta}))$ denotes the model probabilities of each expert $e$, $\sigma (\cdot)$ denotes the softmax activation function, and $p(\cdot \mid \bm{v}_j(x) \ell; \bm{\theta}_{\beta})) : \mathbb{R}^{d} \mapsto \mathbb{R}^{d_e}$ denotes a gating function conditioned on the input $\bm{v}(x)$ and the PDE label $\ell$. The gating function is parameterized by its own parameters $\bm{\theta}_{\beta}$. The term $\phi_{j}^{e}$ denotes the parameters of the $e^{th}$ expert in the $j^{th}$ hidden layer. In this work, the transformation P, Q, and $g$ are designed as $1 \times 1$ convolution neural networks. The gating function is modeled as a four-layered feedforward network with the mish activation function \cite{misra2019mish}. 
To parameterize the kernel $k^{e}_{\phi_j}(\bm{a}(x),\bm{a}(\xi),x,\xi)$ in Eq. \eqref{eq:integral}, we drop the terms $\bm{a}(\cdot)$ and modify the kernel as $k^{e}_{\phi_j}(x-\xi)$. This modification allows us to express the kernel integral operator $\mathcal{K}(\bm{a}(x); \phi)$ as a convolution integral between the network kernel $k^{e}_{\phi_j}$ and input $\bm{v}_j(x)$, given as,
\begin{equation}\label{eq:convolve}
    \left(\mathcal{K}(\phi_j \in \bm{\theta}) \bm{v}_{j}\right)(x) = \sum_{e=1}^{d_e} \left[ \bm{\beta}^{e}_{j} \cdot \int_{\mathbb{R}^{d}} k^{e}_{\phi_j} \left(x- \xi \right) \bm{v}_{j}(\xi) \mathrm{d} \xi \right]; \;\; x \in \Omega^d, \;\; j \in [1,h].
\end{equation}
where $k^{e}_{\phi_j}$ is the kernel associated with the $e^{th}$ expert in the $j^{th}$ expert integral block, parameterized by $\phi_{j}^{e} \in \bm{\theta}$ which denotes the model parameters of the $e^{th}$ expert in the $j^{th}$ hidden iteration. 
With the undecimated wavelet transform, the convolution integral in Eq. \eqref{eq:convolve} becomes element-wise multiplication between input $\bm{v}_j$ and kernel weights $k^{e}_{\phi_j}$ in the wavelet domain. The process involves wavelet decomposition of the input $\bm{v}_j$, followed by convolution between the decomposed input and the local kernel $k^{e}_{\phi_j}$. The variation in the local behavior of each local wavelet expert is introduced by using different wavelet basis functions for different local wavelet experts. 
We define the wavelet basis functions as $\psi_{e} \in L^2(\mathbb{R}^n)$. Further, denoting the wavelet and inverse wavelet transforms as $W_{\psi_{e}}(\cdot)$ and $W_{\psi_{e}}^{-1}(\cdot)$ with wavelet basis $\psi_{e}$, respectively, we express Eq. \eqref{eq:convolve} in terms of the convolution operation in the wavelet domain as,
\begin{equation}\label{eq:convolve_tensor}
    \left(\mathcal{K}(\phi_j \in \bm{\theta}) \bm{v}_{j}\right)(x) = \sum_{e=1}^{d_e} \left[ \bm{\beta}_{e} \cdot \left\{ W_{\psi_{e}}^{-1} \left( \kappa^{e}_{\phi_j} \odot W_{\psi_{e}}\left(\bm{v}_{j} ;s,r \right) \right) (x) \right\} \right]; \;\; x \in \Omega^d, \;\; j \in [1,h].
\end{equation}
where $\kappa_{\phi_j}^{e} := W_{\psi_e}(k_{\phi_j}^{e})$ denotes the network kernel of the $e^{th}$ expert directly defined in the wavelet domain. $[\cdot]$ denotes the Schur product that provides the channel-level probabilities, allowing the channel-level ensembling of the outputs of the local kernel integral $\int_{\mathbb{R}^{d}} k^{e}_{\phi_j} \left(x- \xi \right) \bm{v}_{j}(\xi) \mathrm{d} \xi$, $e = 1,\ldots, d_e$.
The undecimated DWT of the input signal $\bm{v}_j(x)$ provides the detailed and approximation components $\mathbb{D}_v$ and $\mathbb{A}_v$. After $s$ level of compression, the support width of the detailed and approximate wavelet coefficients can be estimated as $d_{\omega} = d/2^s + 2(d_{\psi}-1)$, where $d$ is the signal length, $s$ is the compression level, and $d_{\psi}$ is the dimension of the vanishing moment of the undertaken wavelet basis.
An $s$ level wavelet compression's of the input $\bm{v}_{t} \in \mathbb{R}^{d \times d_{v}}$ provides $\mathbb{R}^{d_{\omega} \times d_{v}}$ dimensional tensors for $\mathbb{D}_v$ and $\mathbb{A}_v$. Since the highest compression level contains the most significant information, we parameterize the kernels $k_{\phi_j}^{e}$ in the highest compression layer $s$. To parameterize the network, we define $\kappa^{e}_{\phi_j} \in \mathbb{R}^{d_{\omega} \times d_{v} \times d_{v}}$ dimensional kernels for the local wavelet expert. 
Through the direct parameterization of the local wavelet kernels $\kappa^{e}_{\phi_j}$ in the wavelet domain, we express the expert kernel integral operator in \eqref{eq:convolve_tensor} as follows,
\begin{equation}\label{eq:convolution_final}
    \left(\mathcal{K}(\phi_j) \bm{v}_{j}\right)(x) = \sum_{e=1}^{d_e} \left[ \bm{\beta}_{e} \cdot \left\{ W_{\psi_{e}}^{-1} \left( \sum_{i_{1}=1}^{d_{v}} \left(\kappa^{e}_{\phi_j}\right)_{i_0, i_1, i_2} W_{\psi_{e}} \Bigl(\bm{v}_{j}; s, r\Bigr)_{i_0, i_1} \right) (x) \right\} \right] \;\; x \in \Omega^d, \;\; j \in [1,h].
\end{equation}
Overall, the parameterization of the network against the input $\bm{v}_j(x)$ is done using two types of expert kernels, denoted as $\kappa^{e}_{\phi_j, \mathbb{I}=\mathbb{D}}$ for detailed coefficients and $\kappa^{e}_{\phi_j, \mathbb{I}=\mathbb{A}}$ for approximate coefficients. Using these kernels, we learn the solution operator $\mathcal{D}$. The algorithmic implementation is provided in Algorithm \ref{algo_expert}.

\subsection{Mixture gates}
The mixture gate provides NCWNO with the probabilities $\bm{\beta} \in \mathbb{R}^{d_e}$ for channel-level averaging of the local wavelet expert outputs. Each iteration $q_j : \mathbb{R}^{d_v} \mapsto \mathbb{R}^{d_v}$ in the NCWNO is accompanied by an independent gate. The gates are defined as a function of two inputs. In the discretized domain $\Omega \in \mathbb{R}^d$, the first input is the context function $\bm{v}_j \in \mathbb{R}^{d \times d_v}$, and the second input is the task label $\ell \in \mathbb{Z}^{+}$. Given the set $\{\bm{v}_j, \ell \}$, the probabilities $\bm{\beta} \in \mathbb{R}^{d_e \times d_v}$, predicting the importance of an expert $e$ across $d_v$ can be estimated as,
\begin{equation}\label{eq:probability}
    \bm{\beta}_{i_0,i_1} = \frac{\operatorname{exp} \left[p_{\beta}(e \mid \bm{v}_{j}, {\mathcal{E}}_{d_e}(\ell); \varphi_e \in \bm{\theta}_{\beta})_{i_0,i_1}\right]}{\sum^{d_e}_{i_0=1} \operatorname{exp} \left[p_{\beta}({i_0} \mid \bm{v}_{j}, {\mathcal{E}}_{d_e}(\ell); \varphi_{i_0} \in \bm{\theta}_{\beta})_{i_0,i_1} \right]}; \;\; i_1 \in d_v, \; e \in [1,d_e],
\end{equation}
where ${\mathcal{E}}_{d_e}(\ell)$ is a linear encoding of $\ell$, and $p_{\beta}(e \mid \bm{v}_j, {\mathcal{E}}_{d_e}(\ell); \varphi_e) \in \mathbb{R}^{d_e \times d_v}$ is a gating function parameterized by the model parameters $\varphi_e \in \bm{\theta}_{\beta}$, conditioned on the input $\bm{v}_j$ and the task label $\ell$. We design the gating function $p_{\beta}(\cdot)$ as a feedforward network with mish activation in the hidden layers and softmax activation at the network's output. The encoder ${\mathcal{E}}_{d_e}(\cdot)$ is designed a three-layered linear feed-forward layer.

\subsection{Combinatorial transfer learning on new parametric PDEs using NCWNO}
The combinatorial transfer learning involves initializing and training the NCWNO parameters $\bm{\theta}$ along with the parameters of the gating function $\bm{\theta}_{\beta}$ for $m$ number of initial PDEs $\{\mathcal{N}_{0}, \ldots, \mathcal{N}_{m} \}$. Once the universal solution operator $\mathcal{D}^{\mathcal{N}_{0:m}}_{\bm{\theta}}$ of the set $\{\mathcal{N}_{0}, \ldots, \mathcal{N}_{m} \}$ is learned, the combinatorial transfer learning is achieved by fine-tuning the parameters of the gate function $\bm{\theta}_{\beta}$, while making the parameters of the foundation model $\bm{\theta}$ stationary.
Given a set of $m$ new PDEs, the solution operators $\mathcal{D}^{\mathcal{N}_{0:m} \cup \tau}_{\bm{\theta}}$ for $\tau \in \{\mathcal{N}^{*}_1, \ldots, \mathcal{N}^{*}_m \}$ is learned by treating the PDE $\mathcal{N}^{*}_{\tau}$ to be an independent task. At each new task level $\tau$, the trained parameters $\bm{\theta}_{\beta}^{\tau}$ of the gate function is stored in a pool of semantic memory, allowing to perform accurate predictions in old tasks. By getting rid of the need for saving large expert models, the NCWNO achieves both data and resource efficiency. Besides continual learning of parametric PDEs, the ability to acquire and assimilate new knowledge of physical systems through efficient semantic memory allows the NCWNO to perform zero-shot predictions on old tasks without catastrophic forgetting and rehearsal \cite{jeeveswaran2023birt} of the network parameters on old PDEs. 
The algorithmic implementation is provided in Algorithm \ref{algo_wno}.

\subsection{Complexity of the expert wavelet integral layer}
Designing the linear transformation $g(\cdot)$ as a $1 \times 1$ convolution for an input with spatial dimension $\mathbb{R}^{d}$ has the complexity of $\mathcal{O}(d)$. The undecimated DWT has a time complexity of $\mathcal{O}(d)$. An $s$ level of wavelet compression of the $d$-dimensional input results in wavelet coefficients of size $\mathbb{R}^{d_w}$. Wavelet domain convolution of the wavelet coefficients and weight tensor takes $\mathcal{O}(d_w)$ time. Since $d_w < d$, the time complexity of the expert wavelet integral layer is $\mathcal{O}(d)$.

\subsection{Implementation and hyperparameter setting}
The foundation NCWNO model consists of four expert wavelet integral blocks. Each expert wavelet integral block consists of 10 local wavelet experts. The first 10 wavelet basis functions from the Daubechies family are chosen as the wavelet basses \cite{daubechies1992ten}. The mish activation function \cite{misra2019mish} is employed. The transformations P and Q are designed as a single layer $1 \times 1$ convolution neural network (CNN) with 64 and 128 channels. The linear skip connection layer $g(\cdot)$ is designed as $1 \times 1$ CNN with 64 channels. 
For the 1D problems, the gating function $p_{\beta}$ is designed as a 6-layered feed-forward network with mish activation and softmax at the output layer. The ADAM optimizer with an initial learning rate of 0.001 and a weight decay of $10^{-6}$ is utilized. For training the foundation model, a step scheduler with stepsize 20 for a total of 150 epochs and a decay rate of 0.5 is used. The wavelet compression level is considered as 4. During the combinatorial transfer, a total of 50 epochs for each new parametric PDE with a stepsize of 20 is used.
For the 2D problems, the gating function $p_{\beta}$ is composed of a 3-layered CNN and a 4-layered feed-forward network. The CNN layers have a kernel size of 5, a stride of 2, and a mish activation function. The feed-forward layers are activated using the mish activation function within the layers, and a softmax is placed at the output layer. The optimizer parameters are kept the same as the 1D problems. For training the foundation model, a step scheduler with stepsize 20 for a total of 100 epochs and a decay rate of 0.5 is used. The wavelet compression level is considered as 4. During the combinatorial transfer, the same parameters as the 1D problems are used.
All training and testing are performed on a single Nvidia RTX A5000 24GB GPU card.

\subsection{Data sets}

\paragraph{Burgers equation.}\label{sec:1d_burger}
As a first example, we consider the Burgers equation that draws its significance from its use in the modeling of flows in fluid mechanics, gas dynamics, and traffic flow. The Burgers equation is given as, $\partial_{t} \bm{u}(\bm{x}, t)+0.5\partial_{x}\bm{u}^{2}(\bm{x}, t) =\nu \partial_{x x} \bm{u}(\bm{x}, t)$ with $\bm{u}(\bm{x}, 0) =\bm{u}_{0}(\bm{x})$, where $\nu \in \mathbb{R}^{+}$ is the viscosity of the flow and $\bm{u}_{0}(\bm{x}) \in \mathbb{R}$ is the initial condition. The synthetic data are generated for different initial conditions $\bm{u}_{0}(\bm{x})$, simulated as Gaussian random fields (GRF) with the radial basis function (RBF) kernel ${K}(x,x\prime) = \sigma^2 \operatorname{exp}(-\|x-x\prime\|^2/2\ell^2)$ and the viscosity is taken as $\nu=10^{-3}$. We adopt a periodic boundary condition $\bm{u}(\bm{x}=0, t) = \bm{u}(\bm{x}=1, t)$. In the 2D setting, the domain $\Omega=(0,1)^2$ is discretized into $64\times64$ grid, and in the 1D setting the domain $\Omega=(0,1)$ is discretized into 256 grid. In 2D, different $\bm{u}_{0}(\bm{x})$ are generated using the kernel parameters $\sigma=0.1$ and $\ell=0.3$, where as in 1D, the parameters are taken as $\sigma=0.1$ and $\ell=0.1$. In both 1D and 2D settings, the second-order FDM is used in space discretization and Euler's scheme for time forwarding, where we utilize a sampling frequency of 1000Hz to solve the system, while the data are recorded at every $t$=0.01s. 
\paragraph{Wave equation.}\label{sec:wave}
The wave equation is a second-order linear hyperbolic PDE used in describing traveling waves in solids and fluids. The 1D wave equation is given as $\partial_{tt} u(x, t) =\nu \Delta u(x, t)$ with $u(x, 0) = u_{0}(x)$, where $\nu \in \mathbb{R}^{+}$ is a non-negative real coefficient denoting the speed of a wave. We use the reflective boundary condition $\partial_x u(x=0, t) = \partial_x u(x=1, t) = 0$ for simulating the equation. For data generation, we discretize the domain $\Omega=(0,1)$ into a 256 grid. We consider the speed of the wave to be $\nu=0.1$. Different initial conditions $u_{0}(x)$ are simulated from the RBF kernel using the kernel parameters $\sigma=0.1$ and $\ell=0.1$. In the 1D setting, we use second-order FDM in space and Euler's scheme in time with a sampling frequency of 1000Hz, while the data are recorded at every $t$=0.01s. 
\paragraph{Advection equation.}\label{sec:advection}
The advection equation, which primarily describes the solution of a scalar under a known velocity field, follows the following PDE, $\partial_t u(\bm{x},t) + \alpha \partial_x u(\bm{x},t) = 0$ with $u(\bm{x}, 0) =u_{0}(\bm{x})$, where $\alpha \in \mathbb{R}^{+}$ is the velocity of motion. A periodic boundary condition $u(\bm{x}=0, t) = u(\bm{x}=1, t)$ is used for simulating the data.
The training data for the 2D case is generated on $\Omega \in (0,1)^2$ with a spatial resolution $64 \times 64$. Different initial conditions are generated as a GRF using the RBF kernel with parameters $\sigma=0.1$ and $\ell=0.3$. The velocity of motion is adopted as $\alpha=0.05$. The second-order FDM is utilized for solving the system in space, while time forwarding is done using Euler's scheme. A sampling frequency of 1000Hz is used for solving the system, and a sampling frequency of 100Hz is used to record the data. 
Data generation for the 1D advection equation involves solving the above equation on the domain $\Omega = (0,1)$ with a 256 spatial grid. $\alpha$ is taken as 0.01. In the 1D domain, the RBF kernel parameters for different initial conditions are taken as $\sigma=0.1$ and $\ell=0.25$. The same solver with identical specifications is used to generate and record the data. 
\paragraph{Heat equation.}\label{sec:heat}
The heat equation is a parabolic linear partial differential equation often used in explaining the conduction of heat through a medium, given as $\partial_{t} u(\bm{x}, t) = \alpha \Delta u(\bm{x}, t)$, where $\alpha \in \mathbb{R}^{+}$ is a strictly positive quantity called the thermal diffusivity of the medium. 
For the 2D case, we adopt $\alpha = 10^{-3}$ and $\Omega \in(0,1)^2$. For synthetic data generation, different initial conditions are simulated from the RBF kernel using $\sigma=0.1$ and $\ell=0.25$. The equation is solved using the pseudospectral method on $64 \times 64$ grid with a sampling frequency of 1000Hz, however, the training data are recorded at every t=0.1s. 
The training data for the 1D heat equation is generated by solving the above equation in $\Omega=(0,1)$ for $\alpha = 10^{-3}$. The initial conditions are simulated from RBF kernel using $\sigma=0.1$ and $\ell=0.1$. For the 1D setting, we use FDM with a sampling frequency of 1000Hz to solve the system and a sampling frequency of 100Hz to record the data. 
\paragraph{Allen-Cahn equation.}\label{sec:AC}
The Allen–Cahn equation is a second-order nonlinear parabolic partial differential equation that has been extensively used to study various physical problems like crystal growth, image segmentation, and phase separation in the multi-component alloy. The mathematical description of Allen-Cahn follows, $\partial_t u(\bm{x},t) = \epsilon \partial_{xx} u(\bm{x},t) + u(\bm{x},t) - u(\bm{x},t)^3$ with $u(\bm{x},0) = u_0(\bm{x})$, where $\epsilon \in \mathbb{R}^{+}$ is a real positive constant responsible for the speed of diffusion. With a periodic condition $u(\bm{x}=0, t) = u(\bm{x}=1, t)$, the synthetic datasets are generated for different initial conditions $u_{0}(x)$.
In the 2D setting, the domain is taken as $\Omega \in (0,1)^2$, and the speed of diffusion is taken as $\epsilon = 10^{-3}$. The initial conditions are generated on a $64 \times 64$ spatial grid using the RBF kernel with parameters $\sigma=0.1$ and $\ell=0.1$. The system is solved using a spectral finite element method with a sampling frequency of 1000Hz.
For generating training data for the 1D Allen-Cahn equation, the above equation is solved in $\Omega=(0,1)$ with a 256-point spatial grid. The same kernel parameters are considered for generating $u_{0}(x)$ in 1D. A spectral method identical to the 2D case is used to solve the system. In both 1D and 2D settings, a sampling frequency of 1000Hz is used to solve the system, while the training data are recorded at every $t$=0.01s. 
\paragraph{Navier-Stokes equation.}\label{sec:ns}
We consider the 2D incompressible Navier-Stokes (NS) equation, in the vorticity-velocity form $\partial_{t} \omega(\bm{x}, t)+u(\bm{x}, t) \cdot \nabla \omega(\bm{x}, t) =\nu \Delta \omega(\bm{x}, t)+f(\bm{x})$ and $ \nabla \cdot u(\bm{x}, t) =0$. Here $\nu \in \mathbb{R}^{+}$ is the viscosity of the fluid, $f(\bm{x})$ is the source function, $u(\bm{x}, t)$ and $\omega(\bm{x}, t)$ are the unknown velocity and vorticity fields. 
We simulate the NS equation on a 2D square domain $\Omega=(0,1)^2$. For generating training data, we use $\nu = 10^{-3}$ and a constant force field $f(x,y) = 0.1\left(\sin \left(2 \pi\left(x+y\right)\right)+\right. \left.\cos \left(2 \pi\left(x+y\right)\right)\right)$. The initial vorticity field $\omega_0{(x,y)}$ is simulated from a GRF as $\mathcal{N}(0,7^{3 / 2}(-\Delta+49 I)^{-2.5})$. A pseudospectral method is adopted to solve the system. The pressure-Poisson equation is solved in Fourier space and differentiated to obtain the vorticity field, which is forwarded in time using the Crank–Nicolson scheme. Similar to the previous examples, a $\Delta t=10^{-4}$s is for solving the system while the training data are recorded at every $t$=1s. 
\paragraph{Nagumo equation.}\label{sec:nagumo}
As a last example, we consider the non-linear Nagumo equation, which is often used in modeling the transmission of nerve impulses. It is mathematically given as, $\partial_{t} u(\bm{x}, t) =\nu \Delta u(\bm{x}, t)+u(\bm{x},t)(1-u(\bm{x},t))(u(\bm{x},t)-\alpha)$, where $\nu \in \mathbb{R}^{+}$ is the speed of the pulse and $\alpha \in \mathbb{R}$ is some constant. Similar to previous examples, we adopt the periodic boundary condition $u(\bm{x}=0, t) = u(\bm{x}=1, t)$. 
In the 2D setting, we adopt the domain $\Omega \in(0,1)^2$ with the spatial resolution $64\times 64$. Synthetic data for different initial conditions are generated by simulating the initial conditions as GRF with the RBF kernel. The RBF kernel parameters are taken as $\sigma^2=0.1$ and $\ell = 0.3$. We use a Galerkin spectral element method with a sampling frequency of 1000Hz to solve the system for $\nu = 10^{-3}$. For training data, we record time snapshots ar=t every $\Delta t$= 0.01s. 
In the 1D setting, the Nagumo equation is solved on $\Omega=(0,1)$ with 256 spatial grid points. Different initial conditions are simulated from the RBF kernel using $\sigma^2=0.1$ and $\ell = 0.1$. A spectral element method with identical parameters to the 2D case is adopted for solving the system.

\section*{Acknowledgements} 
T. Tripura acknowledges the financial support received from the Ministry of Education (MoE), India, in the form of the Prime Minister's Research Fellowship (PMRF). S. Chakraborty acknowledges the financial support received from Science and Engineering Research Board (SERB) via grant no. SRG/2021/000467, and from the Ministry of Port and Shipping via letter no. ST-14011/74/MT (356529).

\section*{Declarations}


\subsection*{Conflicts of interest} The authors declare that they have no conflict of interest.


\subsection*{Code availability} Upon acceptance, all the source codes to reproduce the results in this study will be made available to the public on GitHub by the corresponding author.

\newpage


\section*{Appendix}

\appendix

\section{Algorithm for local ensembling between local wavelet experts}
Through the direct parameterization of the local wavelet kernels $\kappa^{e}_{\phi_j}$ in the wavelet domain, the expert kernel integral operator can expressed as follows,
\begin{equation}
    \left(\mathcal{K}(\phi_j) \bm{v}_{j}\right)(x) = \sum_{e=1}^{d_e} \left[ \bm{\beta}_{e} \cdot \left\{ W_{\psi_{e}}^{-1} \left( \sum_{i_{1}=1}^{d_{v}} \left(\kappa^{e}_{\phi_j}\right)_{i_0, i_1, i_2} W_{\psi_{e}} \Bigl(\bm{v}_{j}; s, r\Bigr)_{i_0, i_1} \right) (x) \right\} \right] \;\; x \in \Omega^d, \;\; j \in [1,h].
\end{equation}
Overall, the parameterization of the network against the input $\bm{v}_j(x)$ is done using two types of expert kernels, denoted as $\kappa^{e}_{\phi_j, \mathbb{I}=\mathbb{D}}$ for detailed coefficients and $\kappa^{e}_{\phi_j, \mathbb{I}=\mathbb{A}}$ for approximate coefficients. Using these kernels, we learn the solution operator $\mathcal{D}$. An algorithmic illustration of the iteration $q:\mathbb{R}^{d_v} \mapsto \mathbb{R}^{d_v}$ is provided in Algorithm \ref{algo_expert}.
\begin{algorithm}[ht!]
    \caption{Combinatorial wavelet integral layer}\label{algo_expert}
    \begin{algorithmic}[1]
	\Require{Input $\bm{v} \in \mathbb{R}^{d \times d_v}$, task level $\ell$, and wavelet basis set $\{\psi_e, \varphi_e \}_{e=1}^{d_E}$.}
    \State{Initiate the gating network: $p_{\beta}(\cdot ; \bm{\theta}_{\beta})$.}
    \For{expert, $e=1,\ldots,d_E$}
        \State{Initiate the weight tensors: $\kappa_{\phi, \mathbb{I}}^{e} \in \mathbb{R}^{d_{\omega} \times d_v \times d_v}$ for $\mathbb{I} \in \{\mathbb{D}_v,\mathbb{A}_v\}$.}
    \EndFor
    \For{Expert $e=1,\ldots,E$}
        \State{Obtain the probability $\bm{\beta}_e \in \mathbb{R}^{d_v}$.} 
    	\State{Multilevel DWT: $\bm{v}_{\psi_e,\mathbb{I}}(s,r) = W_{\psi_{e}}(\bm{v}(x) ; s,r) \in \mathbb{R}^{d_{\omega} \times d_v}$.} 
        \State{Parameterize expert:  $\hat{\bm{v}}_{\psi_e,\mathbb{I}}(s,r) = \sum_{i_{1}=1}^{d_{v}} \bigl( \kappa^{e}_{\phi, \mathbb{I}} \bigr)_{(i_0, i_1, i_2)} \bigl( \bm{v}_{\psi_e,\mathbb{I}}(s,r) \bigr)_{i_0,i_1}$ for $\mathbb{I} \in \{\mathbb{D},\mathbb{A}\}$.} 
        \State{Multilevel IDWT: $\bm{v}_{j+1}^{\prime}(x) \in \mathbb{R}^{d \times d_v}$ = $W_{\psi_{e}}^{-1}\left( \hat{\bm{v}}_{\psi_e,\mathbb{I}}(s,r) \right)$.} 
    \EndFor
    \State{Output $\bm{v}_{j+1} = \Gamma \bigl(\sum^{E}_{e=1} \bm{\beta}_e \odot \bm{v}_{j+1}^e + (g \bm{v}_j) \bigr)$.} 
    \end{algorithmic}
\end{algorithm}

\section{Algorithm of the Neural Combinatorial Wavelet Neural Operator}
Given a set of $m$ new PDEs, the solution operators $\mathcal{D}^{\mathcal{N}_{0:m} \cup \tau}_{\bm{\theta}}$ for $\tau \in \{\mathcal{N}^{*}_1, \ldots, \mathcal{N}^{*}_m \}$ is learned by treating the PDE $\mathcal{N}^{*}_{\tau}$ to be an independent task. At each new task level $\tau$, the trained parameters $\bm{\theta}_{\beta}^{\tau}$ of the gate function is stored in a pool of semantic memory, allowing to perform accurate predictions in old tasks. By getting rid of the need for saving large expert models, the NCWNO achieves both data and resource efficiency. Besides continual learning of parametric PDEs, the ability to acquire and assimilate new knowledge of physical systems through efficient semantic memory allows the NCWNO to perform zero-shot predictions on old tasks without catastrophic forgetting and rehearsal of the network parameters on old PDEs.
An algorithm explaining the steps of NCWNO is provided in Algorithm \ref{algo_wno}.
\begin{algorithm}[ht!]
    \caption{Neural combinatorial wavelet neural operator (NCWNO)}\label{algo_wno}
    \begin{algorithmic}[1]
	\Require{Initial tasks set $\{\mathcal{N}_{0}, \ldots, \mathcal{N}_{m} \}$ , input-output pairs $[ \{ a_i^{\tau}(x), u_i^{\tau}(x) \}_{i=1}^{N} ]_{\tau=1}^{m}$, PDE labels: $\{\ell_{\tau}\}_{\tau=1}^{m}$, coordinates $x \in \Omega$, wavelet basis set $\{\psi_e, \varphi_e \}_{e=1}^{E}$, network hyperparameters.}
        \State{Train the foundation model: $\mathcal{D}^{\mathcal{N}_{0:m}}_{\bm{\theta}}: \{\mathcal{A}_0, \ldots, \mathcal{A}_m \} \mapsto \{\mathcal{U}_0, \ldots, \mathcal{U}_m \}$.}
        \For{every new PDE, $\tau \in \{\mathcal{N}^{*}_1, \ldots, \mathcal{N}^{*}_m \}$}
        \For{epoch = $1,\ldots,$epochs}
        \State{Get the input-output pairs $\{a^{\tau}_i(x),u^{\tau}_i(x) \}_{i=1}^{N}$.}
        \State{Uplift inputs: $v_{0}^{\tau}(x) \in \mathbb{R}^{d\times d_v}$ = P$(a^{\tau}(x))$.}
            \While{$j \le h$}  
        	\State{$v_{j+1}^{\tau} \in \mathbb{R}^{d \times d_v}$ = $q_{\bm{\theta}}(v_{j}^{\tau})$.} 
            \EndWhile
        \State{Compute the loss: $\mathcal{L} \left(u^{\tau}(x), {\rm{Q}}(v_{h}^{\tau}(x)) \right)$.}
        \State{Update the gating network parameters: $\bm{\theta}_{\beta} = \bm{\theta}_{\beta} - \alpha \nabla_{\bm{\theta}_{\beta}} \mathcal{L} \left(u_{\tau}, {\rm{Q}}(v_{h}^{\tau}(x)) \right)$.}
        \EndFor
        \State{\textbf{Output}: Solution operator $\mathcal{D}^{\mathcal{N}_{0:m} \cup \mathcal{N}_{\tau}}_{\bm{\theta}}: \{\mathcal{A}_0, \ldots, \mathcal{A}_m \bigcup \mathcal{A}_{\tau}\} \mapsto \{\mathcal{U}_0, \ldots, \mathcal{U}_m \bigcup \mathcal{U}_{\tau} \}$.}
        \EndFor
    \end{algorithmic}
\end{algorithm}

\section{Bench-marking on single parametric PDEs from literature}\label{sec:benchmark}
In this section, we train the foundation model on a single dataset to learn the task-specific solution operators of parametric PDEs from the literature. We consider two 1D time-independent PDEs, two 1D time-dependent PDEs, two 2D time-independent PDEs, and two 2D time-dependent PDEs from literature, which in a group represent various physical phenomena and applications in science and engineering. We use the relative $L^2$ norm of the prediction error as a performance measure. The measure is defined as $\epsilon = \|\bm{u} - \bm{u}^{*}\| / \|\bm{u}\|$, with $\bm{u}^{*}$ and $\bm{u}$ are the predictions and ground truth, respectively. The test sample size in each example is 100.

\subsection{1D Burgers equation}
\begin{figure}[ht!]
    \centering
    \includegraphics[width=0.8\textwidth]{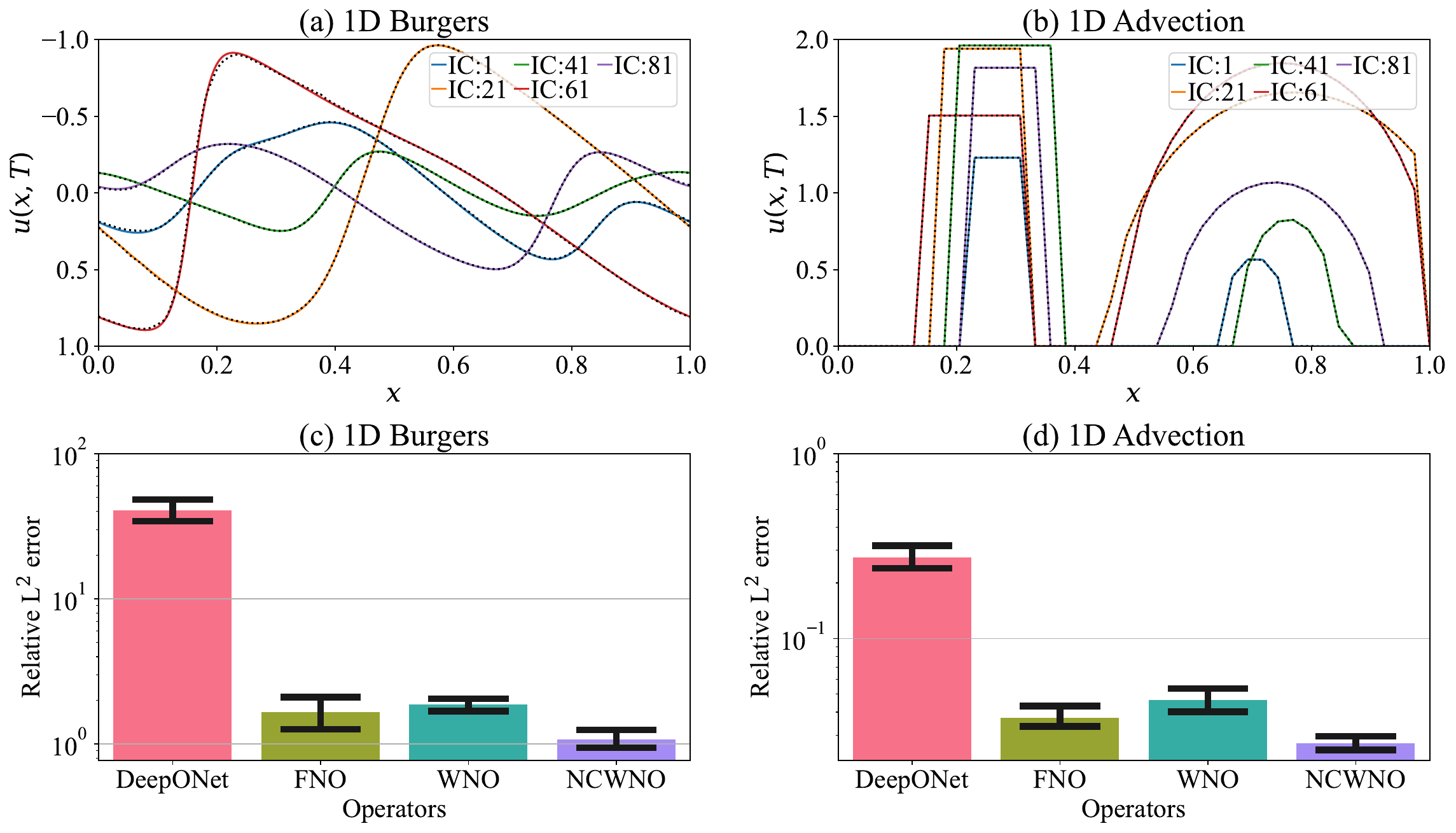}
    \caption{\textbf{Learning solution operator of 1D Burgers' and 1D Wave advection equation}. (a) Prediction for Burgers' equation using NCWNO on a spatial grid of 1024. (b) Prediction for Wave advection equation on a spatial grid for 40. In both (a) and (b), the predictions of NCWNO for five different representative initial conditions are shown, where the solid lines denote the ground truth, and the dotted lines denote the predictions from NCWNO. In both cases, the given initial conditions are directly mapped to the final solution at $T$=1s. (c), (d) $L^2$ relative error in the predictions averaged over 100 different initial conditions for Burgers' and Wave advection equations, respectively.}
    \label{fig_burger1d}
\end{figure}

As a first example, we consider the 1-D Burgers equation from \cite{li2020fourier}. The Burgers equation is given as,
\begin{equation}\label{eq:burgers}
    \begin{aligned}
    \partial_{t} u(x, t)+0.5\partial_{x}u^{2}(x, t) &=\nu \partial_{x x} u(x, t), & & x \in(0,1), t \in[0,T] \\
    u(x=0, t) &= u(x=1, t), & & t \in[0, T] \\
    u(x, 0) &=u_{0}(x), & & x \in(0,1).
    \end{aligned}
\end{equation}
where $\nu \in \mathbb{R}^{+}$ is the viscosity of the flow and $u_{0}(x)$ is the initial condition. For the 1D Burgers equation, the synthetic data are generated for different initial conditions. The initial conditions are generated using a Gaussian random field as $u_0(x) \sim \mathcal{N}(0,625(-\Delta+25 I)^{-2})$ using $\nu$=0.1. The dataset has a spatial resolution of 8192; however, we perform the evaluation on a spatial resolution of 1024. We aim to learn the time-independent operator, $\mathcal{D}: u_{0}(x) \mapsto u(x,1)$, which directly maps the initial condition to the final solution.
\textbf{Results}: The prediction results for five different representative initial conditions are illustrated in Fig. \ref{fig_burger1d}(a). It can be seen that the prediction results completely overlap the true solutions. Further, in Fig. \ref{fig_burger1d}(c), the statistics of the predictive error over 100 different initial conditions are illustrated. The predictive performance of the proposed NCWNO is also compared against existing popular operator learning methods like the DeepONet, FNO, and WNO. From the error statistics, averaged over 100 different initial conditions, it is evident that the NCWNO provides the most accurate predictions for the 1D Burgers equation.
\subsection{1D wave advection equation}
The second example is the wave advection equation, taken from \cite{lu2022comprehensive}, which primarily describes the solution of a scalar under a known velocity field. The synthetic data for the advection equation in the 1D domain is generated using the following equation,
\begin{equation}\label{eq:advection}
    \begin{aligned}
    \partial_t u({x},t) + \nu \partial_x u({x},t) &= 0, & & {x} \in (0,1)^2, t\in [0,T] \\
    u({x}=0, t) &= u({x}=1, t), & & t \in[0, T] \\
    u({x}, 0) &=u_{0}({x}), & & {x} \in(0,1)^2.
    \end{aligned}
\end{equation}
where $\nu \in \mathbb{R}^{+}$ is the velocity of motion and $u_{0}(x)$ is the initial condition. The dataset is generated for different initial conditions. With $\omega$, $h$, and $c$ representing the width, height, and center of the square wave, respectively, the initial conditions are simulated as,
\begin{equation}\label{wave_init}
        u(x,0) = h1_{ \left\{c- 0.5\omega, c+0.5\omega \right\} } + \sqrt{ \max(h^2 - (a(x-c))^2,0) }.
\end{equation}
The values of $\{c, \omega, h\}$ are randomly chosen using a uniform distribution with the parameters $[0.3, 0.7] \times [0.3, 0.6] \times [1, 2]$. With $\nu$=1 and $\Delta t$=0.025s, the solutions are generated on a spatial grid of 40. The aim is to learn the time-independent operator $\mathcal{D}: u_{0}(x) \mapsto u(x,1)$, which maps the initial conditions to the final time step $T$=1.
\textbf{Results}: The prediction results of the 1D wave advection equation for five representative initial conditions are illustrated in Fig. \ref{fig_burger1d}(b). Similar to the previous example, it can be observed that the predictions made by the proposed NCWNO completely overlap the true solutions. From the predictive error plots in Fig. \ref{fig_burger1d}(d), it is further confirmed that the proposed NCWNO provides the least prediction error.

\begin{figure}[ht!]
	\centering
	\includegraphics[width=\textwidth]{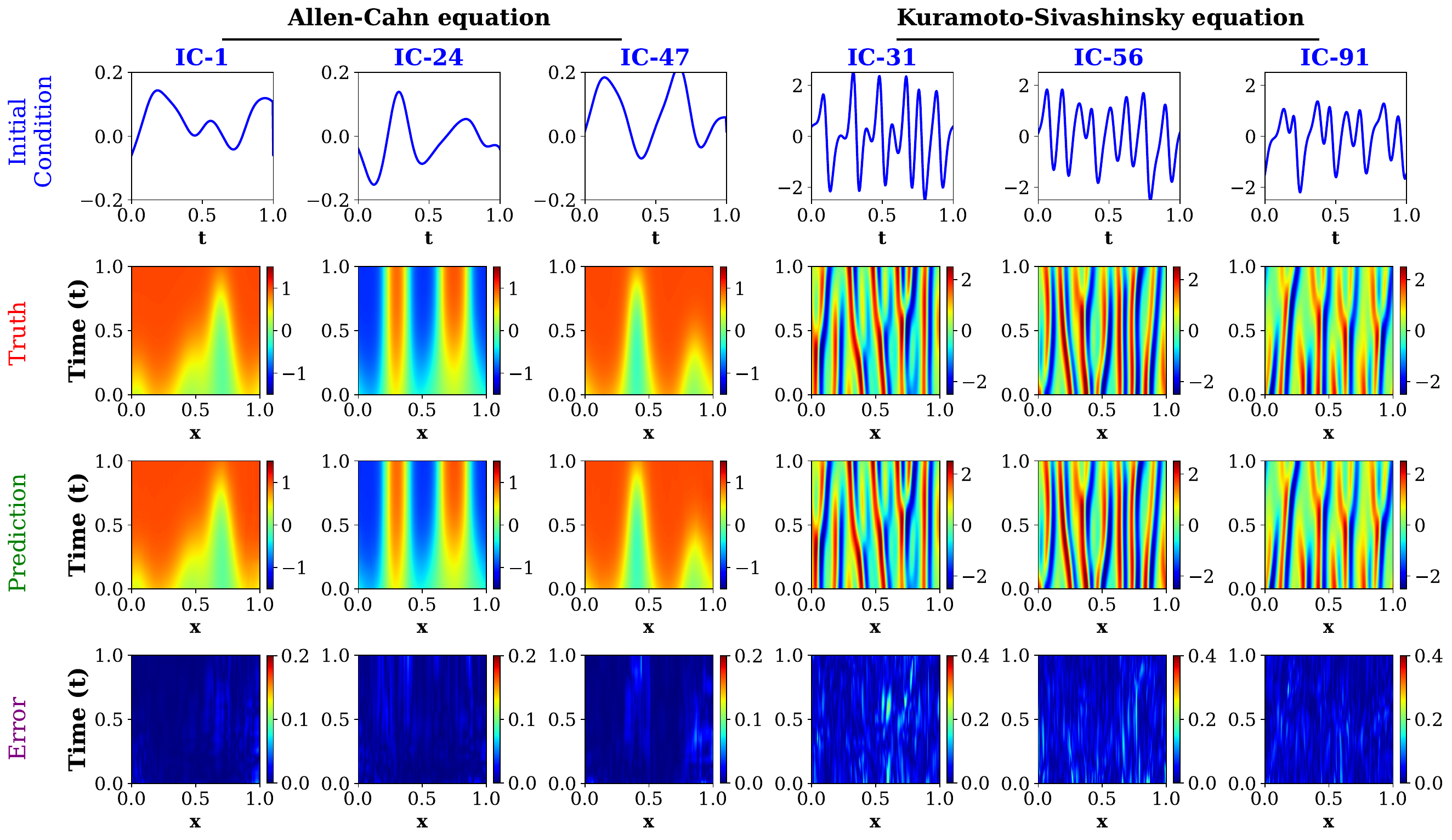}
	\caption{\textbf{Learning solution operator of 1D Allen-Cahn reactor-diffusion equation and 1D chaotic Kuramoto-Sivashinsky equation}. The predictions from NCWNO for three different representative initial conditions are shown. In both examples, the aim is to learn the solution operator $\mathcal{D}$ that maps the first 10 time steps $u_{D\times[0,10]}$ to the next twenty time steps $u_{D\times[11,30]}$ for given unseen initial conditions. In both examples, the spatial domain is fixed at 257.}
	\label{fig_ac1d}
\end{figure}
\subsection{1D Allen-Cahn equation}
\begin{figure}[ht!]
    \centering
    \includegraphics[width=\textwidth]{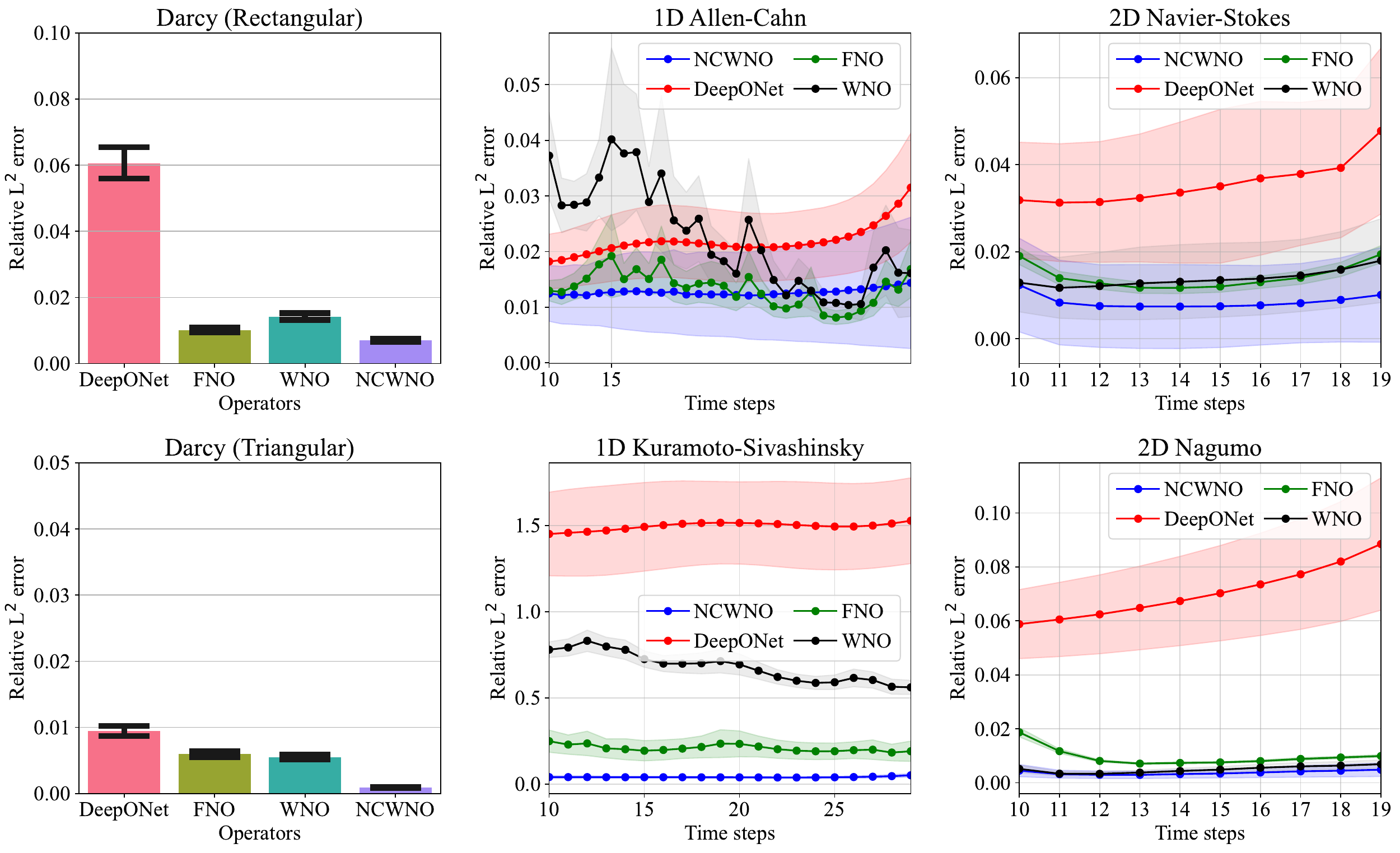}
    \caption{Prediction error between the ground truth and the NCWNO averaged over a hundred different initial conditions. The height of the rectangles on the left and the solid lines in the middle and right indicate the mean $L^2$ error. The error bars and the shaded region show the 95\% confidence error bound.}
    \label{fig:error_benchmark}
\end{figure}

As a third problem, we consider the Allen–Cahn equation from \cite{tripura2023wavelet}. The synthetic data for the Allen-Cahn equation is generated from the following equation,
\begin{equation}\label{eq:allencahn}
    \begin{aligned}
    \partial_t u({x},t) &= \epsilon \partial_{xx} u({x},t) + u({x},t) - u({x},t)^3, & & {x} \in (0,1)^2, t\in [0,T] \\
    u({x}=0, t) &= u({x}=1, t), & & t \in[0, T] \\
    u({x},0) &= u_0({x}) & & {x} \in (0,1)^2
    \end{aligned}
\end{equation}
where $\epsilon \in \mathbb{R}^{+}$ is a real positive constant responsible for the speed of diffusion. The initial conditions are simulated from the Gaussian Random Field using the radial basis function (RBF) kernel, $\mathcal{K}(x,x\prime) = \sigma^2 \operatorname{exp}(-\|x-x\prime\|^2/2\ell^2)$, where the kernel parameters are taken as $\sigma=0.1$ and $\ell=0.15$. The aim here is to learn the time-dependent operator $\mathcal{D}:u_{(0,1)\times[0,10]} \mapsto u_{(0,1)\times[11,30]}$, that maps the solution at first ten-time steps to next thirty time steps. With $\epsilon = 1 \times 10^{-3}$ and $\Delta t$=0.01s, the solutions are generated on a spatial grid of 257.
\textbf{Results}: The time-marching predictions made by the learned solution operator for the 1D Allen-Cahn equation for three different representative test samples are illustrated in Fig. \ref{fig_ac1d}. It can be observed that the predictions made by the proposed NCWNO closely mimic the true solutions. The time evolution of the prediction error is illustrated in Fig. \ref{fig:error_benchmark}, where it can be seen that the prediction error of NCWNO is the least among all the methods and maintains a stable profile throughout the prediction time window, whereas other methods struggle to provide consistent accuracy. This indicates that the NCWNO can also accurately predict time evolutions of the dynamics of underlying systems. 
\subsection{1D Kuramoto-Sivashinsky equation}
The Kuramoto-Sivashinsky (KS) equation is a fourth-order nonlinear partial differential equation \cite{geneva2020modeling}, which is used in modeling several physical systems such as chemical phase turbulence, diffusive–thermal instabilities in a laminar flame, etc. It is widely known for its chaotic behavior when the size of the periodic domain is sufficiently large. Due to its chaotic behavior, predicting the solutions of the KS equation becomes challenging. The KS equation is given as,
\begin{equation}\label{eq:kuromoto}
    \begin{aligned}
    \partial_t u(x,t) &= \left( - u(x,t) \partial_x - \partial_{xx} - \nu \partial_{xxxx} \right) u(x,t), & & x \in (0,22\pi), t\in [0,T] \\
    u(x=0,t) &= u(x=22\pi,t), & & x \in \partial \Omega, t\in [0,T] \\
    u(x,0) &= u_0(x) & & x \in (0,22\pi)
    \end{aligned}
\end{equation}
where $\nu \in \mathbb{R}^{+}$ is referred to as the hyper-viscosity. The synthetic data for the 1D Kuramoto-Sivashinsky (KS) equation is generated by simulating the initial conditions from the Gaussian Random Field using the radial basis function (RBF) kernel, $\mathcal{K}(x,x\prime) = \sigma^2 \operatorname{exp}(-\|x-x\prime\|^2/2\ell^2)$, where the kernel parameters are taken as $\sigma=0.1$ and $\ell=0.1$. The aim here is to learn the time-dependent operator $\mathcal{D}:u_{(0,22\pi)\times[0,10]} \mapsto u_{(0,22\pi)\times[11,30]}$, which maps the solutions at first ten-time steps to next thirty steps. For $\nu = 1$, the solutions are generated on a spatial grid of 257 using $\Delta t$=0.01s.
\textbf{Results}: The predictions from the learned time-dependent solution operator for three different representative test samples are illustrated in Fig. \ref{fig_ac1d}. It is evident that for a slight change in the initial condition, the system becomes a spatiotemporally chaotic attractor. However, the predicted solutions from the NCWNO almost accurately track the time evolution of the true solution. This further indicates that the NCWNO can also accurately learn the solution operators of a spatiotemporally chaotic system, which might be useful for studying evolutions of complex dynamics for different initial conditions or system parameters. The time evolution of the statistics of prediction error is illustrated in Fig. \ref{fig:error_benchmark}, where we see that the DeepONet and WNO fail entirely to approximate the true solution. Although with approximately 3\% error, the FNO makes a good prediction, the proposed NCWNO provides the least predictive error. 

\subsection{2D Darcy flow equation in a rectangular domain}\label{sec:darcy2d}
\begin{figure}[ht!]
	\centering
	\includegraphics[width=\textwidth]{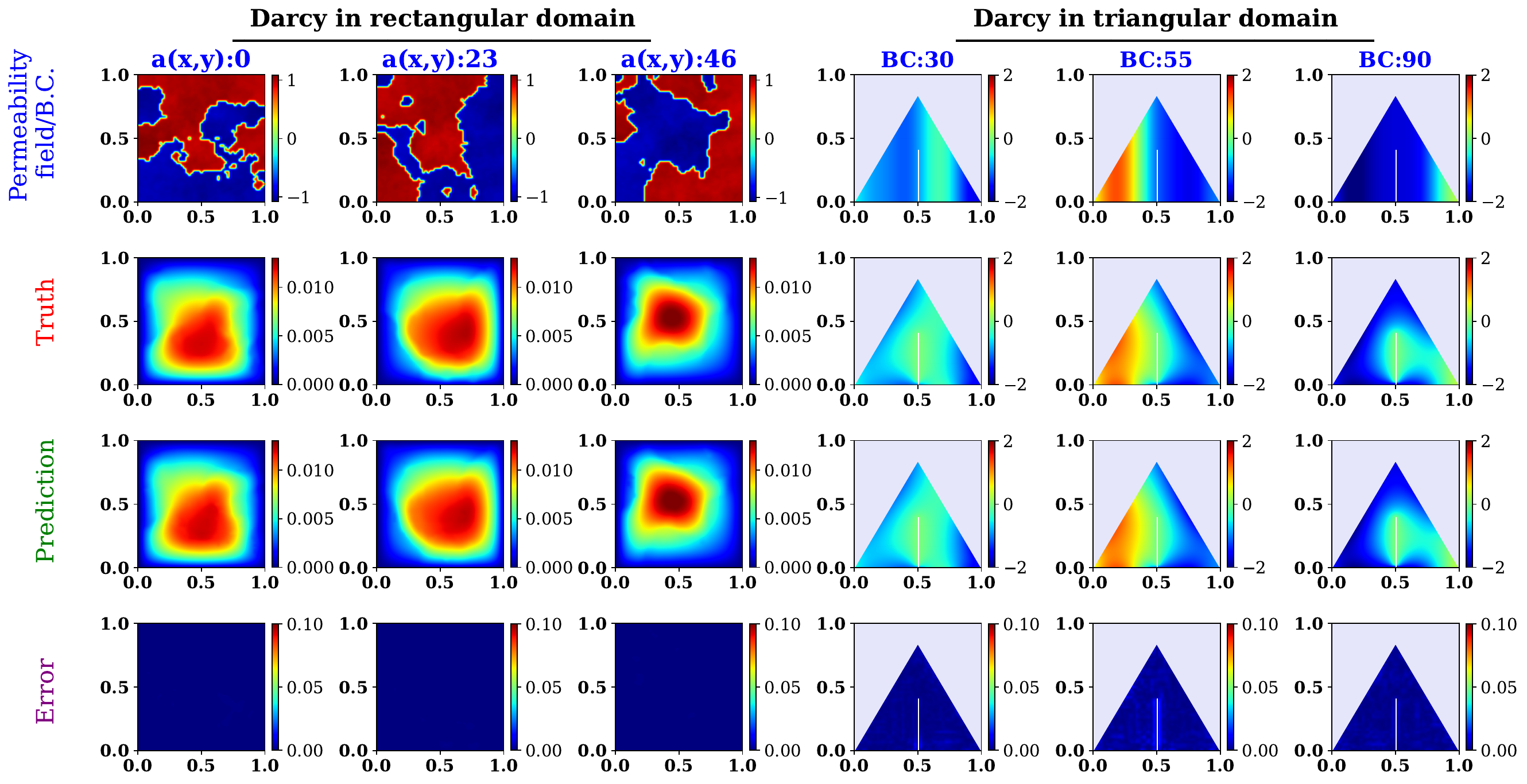}
	\caption{\textbf{Learning solution operator of Darcy flow in the rectangular and triangular domain (with a notch)}. The prediction results are obtained from trained NCWNO. In the first example, the aim is to learn the solution operator $\mathcal{D}$ that maps the given permeability field $a(x,y)$ to pressure field $u(x,y)$. In the second example, the aim is to map the boundary conditions to the solution domain. The solutions corresponding to three representative initial input conditions are shown in each example.}
	\label{fig_darcy2d}
\end{figure}

In the fifth example, we consider the 2D Darcy flow equation from \cite{li2020fourier}, which is a second-order nonlinear elliptic PDE. The Darcy equation follows,
\begin{equation}\label{eq:darcy2d}
    \begin{aligned}
    -\nabla \cdot \left(a(\bm{x}) \nabla u(\bm{x}) \right) &=f(\bm{x}), & & \bm{x} \in(0,1)^2 \\
    u(\bm{x}) &= 0, & & \bm{x} \in \partial (0,1)^2
    \end{aligned}
\end{equation}
where $a(\bm{x})$ is the permeability field, $u(\bm{x})$ is the pressure field, $\nabla u(\bm{x})$ is the pressure gradient, and $f(\bm{x})$ is a source function. $f(x,y)=1$ is considered.
The dataset is generated for different permeability fields, which are generated from a Gaussian random field $u = \psi \mathcal{N}(0,(-\Delta+9\mathbf{I})^{-2})$ with zero Neumann boundary conditions on $\Delta$ (see  \cite{li2020fourier}). The function $\psi: \mathbb{R} \mapsto \mathbb{R}$ pushes the value 12 on the positive part of the real line and the value 3 on the negative line. Although the data are generated on a spatial grid of $421 \times 421$, the training is performed on a spatial resolution of 85 $\times$ 85. The aim is to learn the time-independent operator, $\mathcal{D}: a(x,y) \mapsto u(x,y)$, which maps the permeability fields to the pressure fields.
\textbf{Results}: The predictions of the pressure field for three different representative permeability fields in the rectangular domain are shown in Fig. \ref{fig_darcy2d}. It can be observed that the quality of the predictions is as accurate as the true solutions. The prediction error averaged over 100 different permeability fields are illustrated in Fig. \ref{fig:error_benchmark}. Similar to the previous 1D time-independent and time-dependent examples, we observe that the proposed NCWNO obtains the least prediction error among DeepONet, FNO, and WNO.
\subsection{2D Darcy flow equation with a notch in triangular domain}
In this example, the 2D Darcy problem in Eq. \eqref{eq:darcy2d} is solved on a triangular domain \cite{lu2022comprehensive} with a notch in the flow medium for different boundary conditions. The boundary conditions are generated using the following Gaussian process (GP),
\begin{equation}\label{eq:darcy_notch}
    \begin{aligned}
    u(x) \sim \sigma^2 {\rm{GP}}\left(0, \exp \left(-\frac{(x-x')^2}{2l^2} \right) \right), 
    \end{aligned}
\end{equation}
where $\sigma^2$ and $l$ are the scale and length parameters of the above radial basis function (RBF) kernel. For data generation, the permeability field $a(x,y)$ and the forcing function $f(x,y)$ are set as 0.1 and -1, respectively. The aim here is to learn the operator, $\mathcal{D}: u(x,y)|_{\partial \omega} \mapsto u(x,y)$, which maps the boundary conditions to the pressure field in the entire domain.
\textbf{Results}: The prediction results of the pressure field for three different representative boundary conditions are illustrated in Fig. \ref{fig_darcy2d}. The quality of the predictions and the error plots clearly indicate the robustness of the NCWNO in approximating solutions over complex geometry. From the prediction error statistics in Fig. \ref{fig:error_benchmark}, averaged over 100 different initial conditions, it is further clear that the proposed NCWNO provides the most accurate result among DeepONet, FNO, and WNO.

\subsection{2D Navier-Stokes equation}
We consider the 2D incompressible Navier-Stokes equation from \cite{li2020fourier}, for which we use the following vorticity-velocity equation to generate the synthetic data,
\begin{equation}\label{eq:ns}
    \begin{aligned}
    \partial_{t} \omega(\bm{x}, t)+u(\bm{x}, t) \cdot \nabla \omega(\bm{x}, t) &=\nu \Delta \omega(\bm{x}, t)+f(\bm{x}), & & \bm{x} \in(0,1)^2, t \in[0, T] \\
    \nabla \cdot u(\bm{x}, t) &=0, & & \bm{x} \in(0,1)^2, t \in[0, T] \\
    u(\bm{x}=0, t) &= u(\bm{x}=1, t), & & t \in[0, T] \\
    \omega(\bm{x}, 0) & = \omega_{0}(\bm{x}), & & \bm{x} \in(0,1)^2
    \end{aligned}
\end{equation}
where $\nu \in \mathbb{R}^{+}$ is the viscosity of the fluid, $f(\bm{x})$ is the source function, $u(\bm{x}, t)$ and $\omega(\bm{x}, t)$ are the unknown velocity and vorticity fields. For data generation, $\nu = 10^{-3}$, and $f(x,y) = 0.1\left(\sin \left(2 \pi\left(x+y\right)\right)+\right. \left.\cos \left(2 \pi\left(x+y\right)\right)\right)$ is assumed. The initial vorticity field $\omega_0{(x,y)}$ is simulated from a Gaussian random field $\mathcal{N}(0,7^{3 / 2}(-\Delta+49 I)^{-2.5})$. 
The solutions are generated on a spatial grid of 64 $\times$ 64 using a pseudospectral method (see \cite{li2020fourier} for more details). The time forwarding of the solutions is done using the Crank–Nicolson scheme with a $\Delta t=10^{-4}$s, whereas, for training, the data are recorded at every $t$=1s. The aim is to learn the operator $\mathcal{D}: \omega_{(0,1)^2 \times [0,10]} \mapsto \omega_{(0,1)^2 \times [11,20]}$, i.e., the operator $\mathcal{D}$ that maps the vorticity fields at first ten-time steps to the vorticity fields at next ten time steps.  
\begin{figure}[ht!]
    \centering
    \includegraphics[width=\textwidth]{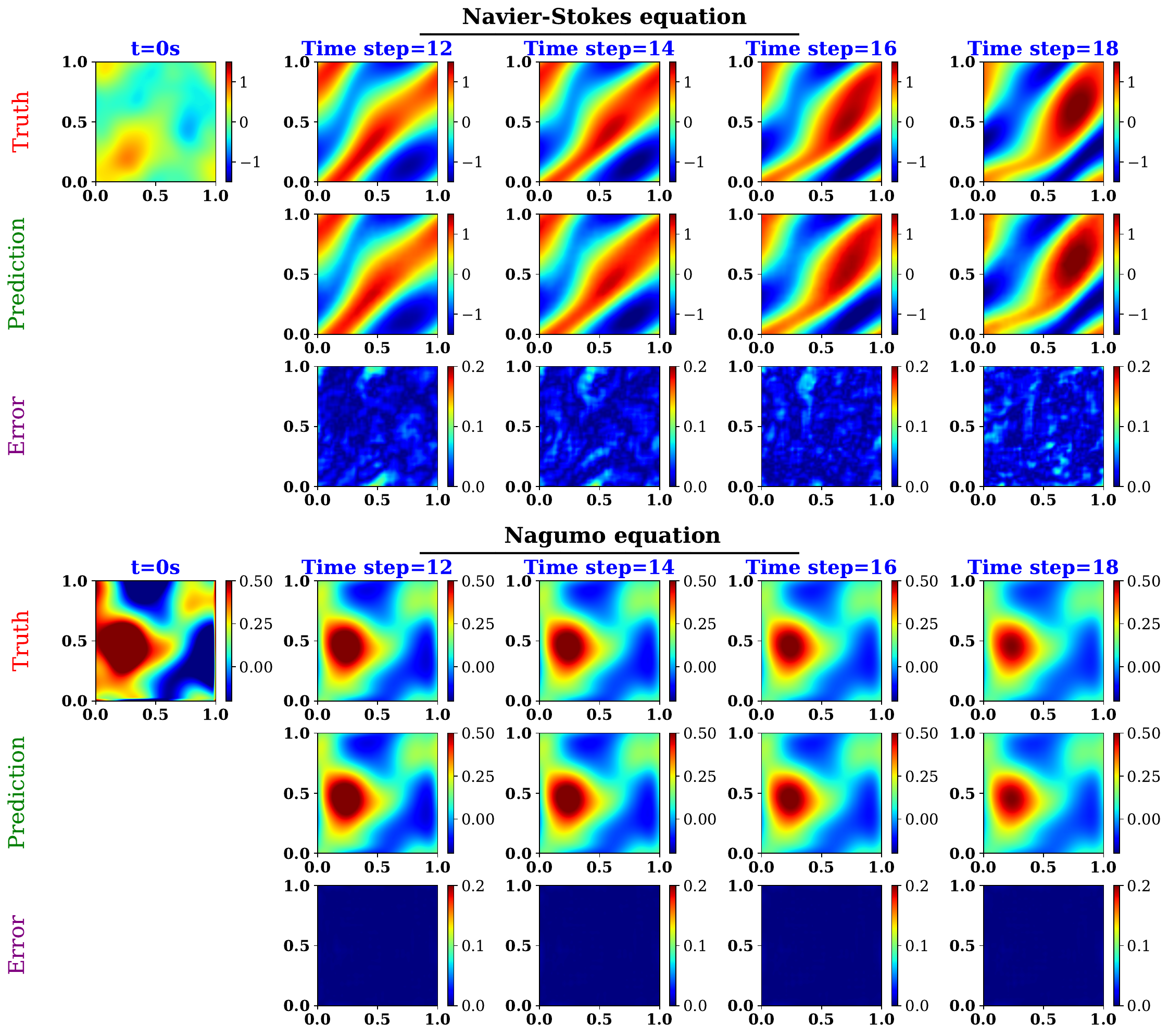}
    \caption{\textbf{Learning time-dependent solution operator of the Navier-Stokes and Nagumo equation with a spatial resolution of $64 \times 64$}. The time evolution of the solutions for a representative initial condition at four time steps is illustrated. The aim in both examples is to learn the solution operator $D$, which maps the solutions $u_{(0,1)^2 \times [0,10]}$ at the initial ten-time steps to the next ten-time steps $u_{(0,1)^2 \times [11,20]}$.}
    \label{fig_NS2dt}
\end{figure}
\textbf{Results}: The time-marching predictions of the vorticity field in the incompressible flow for a representative initial condition are illustrated in Fig. \ref{fig_NS2dt}. Similar to the 1D time-dependent examples, from the error plots, we observe that the prediction results almost accurately mimic the true solutions of the incompressible Navier-Stokes equation. The error confidence bound in Fig. \ref{fig:error_benchmark}, averaged over 100 different initial conditions, further confirms that the proposed NCWNO can accurately learn the solution operator of the 2D time-dependent PDEs.

\subsection{2D Nagumo equation}
Lastly, we consider the non-linear Nagumo equation, which is given by the following 2D PDE,
\begin{equation}\label{eq:ng}
    \begin{aligned}
    \partial_{t} u(\bm{x}, t) &=\nu \Delta u(\bm{x}, t)+u(\bm{x},t)(1-u(\bm{x},t))(u(\bm{x},t)-\alpha), & & \bm{x} \in(0,1)^2, t \in[0, T] \\
    u(\bm{x}=0, t) &= u(\bm{x}=1, t), & & t \in[0, T] \\
    u(\bm{x}, 0) & = u_{0}(\bm{x}), & & \bm{x} \in(0,1)^2 
    \end{aligned}
\end{equation}
where $\nu \in \mathbb{R}^{+}$ is the speed of the pulse and $\alpha \in \mathbb{R}$ is some constant. The dataset is generated for different initial conditions $u_{0}{(x,y)}$ that are simulated from a Gaussian random field with the following Mattern kernel,
\begin{equation}
    \mathcal{K}(x,x\prime) = \sigma^2 \frac{2^{1 - \eta}}{\Gamma(\eta) } \left(\frac{\sqrt{2 \eta}}{\ell} \|x-x\prime\| \right)^{\eta} \hat{K}_{\eta} \left( \frac{\sqrt{2 \eta}}{\ell} \|x-x\prime \| \right),
\end{equation}
where $\hat{K}_{\nu}$ is the modified Bessel function of the second kind. For data generation, we adopt $\eta=10$, $\sigma^2=0.1$, and $\ell = 0.3$. The solutions are generated on a spatial grid of 64 $\times$ 64 using the Galerkin spectral element method scheme with a $\Delta t = 10^{-3}$s, whereas, for training, the data are recorded using $\Delta t$= 0.1s. 
The aim here is to learn the operator $\mathcal{D}: \omega_{(0,1)^2 \times [0,10]} \mapsto \omega_{(0,1)^2 \times [11,20]}$, i.e., the operator $\mathcal{D}$ that maps the vorticity fields at first ten-time steps to the vorticity fields at next ten time steps. 
\textbf{Results}: The time evolution of the solutions of the 2D Nagumo equation for three different representative initial conditions are illustrated in Fig. \ref{fig_NS2dt}. Similar to the previous example, the prediction results from the learned solution operator almost accurately track down the dynamic evolution of the true solution. This can be further confirmed by the error bounds in Fig. \ref{fig:error_benchmark}, where it is evident that the NCWNO obtains the least predictive error among other alternative operator learning schemes.

\section{Network architectures}

\subsection{Architecture of NCWNO}
For continual learning of both 1D and 2D examples, the initial foundation model is trained using four expert wavelet blocks (which denotes the iterations $q_j, j=1,\ldots,h$) are used. Each expert block contains 10 local wavelet experts, where the Daubechies wavelet basis with the first 10 vanishing moments is used. A batch size of 20 with 150 epochs for 1D examples and 100 epochs for 2D examples is used. The encoding transformation P is modeled as a $1 \times 1$ convolution with 64 channels. The decoder transformation Q is modeled as a two-layer $1 \times 1$ convolution with 128 and 1 channels, respectively. The linear skip transformation $g(\cdot)$ is also modeled as $1 \times 1$ convolution layer with 64 channels. A step scheduler with step\_size 20 and decay rate 0.5 is used. The Adam optimizer with weight decay $10^{-3}$ is used. The wavelet compression level is taken as 4 in both 1D and 2D examples. Throughout the network, the "mish" activation function is used to learn the nonlinearity in the functional mapping. 
The gating function is modeled as a dense network with "mish" activation function in the hidden layers and "Softmax" at the output layer. Further details are provided in Table \ref{tab:gate}. 
A summary is provided in Table \ref{tab:ncwno_continual}. 
\begin{table}[ht!]
    \centering
    \caption{Architecture details of gating function $p_{\beta}$. Here, Conv2D denotes the 2D convolution layer with kernel size $k$ and stride $s$. Dense represents the feed-forward network. }
    \label{tab:gate}
    \begin{tabular}{lllllllll}
    \hline
    Continual learning-1D & Network details & \\
    \hline
    Gate network & Dense:[257+6,512,256,128,64,32,10] & \\
    \hline
    Continual learning-2D & \\
    \hline
    Gate network & [64$\times$64,Conv2D(64,k=5,s=2),Conv2D(64,k=5,s=1),Conv2D(64,k=5,s=1),& \\
    &Dense([256+6,128,64,10])] & \\
    \hline
    \end{tabular}
\end{table}
\begin{table}[ht!]
    \centering
    \caption{Architecture details of NCWNO for continual operator learning. }
    \label{tab:ncwno_continual}
    \begin{tabular}{lllllllll}
    \hline
    Examples & P & Q & Compression & Expert blocks/ & Local experts & Channel in local \\
    & & & level, $s$ & global convolution && wavelet convolution \\
    \hline
    1D examples & 64 & [128,1] & 4 & 4/[11,64,64,64,64] & 10 & 64 \\
    \hline
    2D examples & 64 & [128,1] & 4 & 4/[11,64,64,64,64] & 10 & 64 \\
    \hline
    \end{tabular}
\end{table}

For training the NCWNO on single tasks, we use a batch of 10 in Burgers and Darcy with a rectangle domain and 20 in other examples. An initial learning rate of 0.001 is used in all examples. In the Burgers and Darcy with the rectangular domain examples, 500 epochs and a step scheduler with step\_size 25 and rate decay 0.75 are used. In the Darcy with triangular domain and Kuramoto–Sivashinsky examples, 500 epochs and a step scheduler with step\_size 50 and rate decay 0.5 are used. A total of 200 epochs and a step scheduler with step\_size 25 and rate decay 0.5 are used in the Darcy with triangular domain, Allen-Cahn, Navier-Stokes, and Nagumo example.
\begin{table}[ht!]
    \centering
    \caption{Architecture details of NCWNO for task-specific operator learning}
    \label{tab:ncwno}
    \begin{tabular}{lllllllll}
    \hline
    Example & P & Q & Compression & Expert blocks/ & Local experts & Channel in local \\
    &&& level, $s$ & global convolution & & wavelet convolution \\
    \hline
    Burgers & 64 & [128, 1] & 7 & [2, 64, 64, 64, 64] & 10 & 64 \\
    Advection & 64 & [128, 1] & 3 & [2, 64, 64, 64, 64] & 10 & 64  \\
    Allen Cahn & 64 & [128, 1] & 5 & [11, 64, 64, 64, 64] & 10 & 64  \\
    Kuromoto & 40 & [128, 1] & 4 & [11, 40, 40, 40, 40] & 10 & 40 \\
    Darcy (rectangular) & 64 & [128, 1] & 4 & [3, 64, 64, 64, 64] & 10 & 64  \\
    Darcy (triangular) & 64 & [128, 1] & 4 & [3, 64, 64, 64, 64] & 10 & 64  \\
    Navier-Stokes & 30 & [128, 1] & 4 & [12, 30, 30, 30, 30] & 10 & 30  \\
    Nagumo & 30 & [128, 1] & 3 & [12, 30, 30, 30, 30] & 10 & 30  \\
    \hline
    \end{tabular}
\end{table}

\subsection{DeepONet}
For training the DeepONet, the "tanh" activation is used in the Burgers and Advection example. The rectified linear unit (ReLU) is used in other examples. An initial learning rate of 0.001 and $3\times10^{-4}$ are used in the 1D and 2D examples, respectively. and A batch size of 1 is used in all the examples. 500000 epochs are used in the Burgers and Advection example, 250000 epochs are used in the Allen Cahn and Kuramoto–Sivashinsky examples and 100000 epochs are used in other examples. Other network details are provided in Table \ref{tab:deeponet}. 
\begin{table}[ht!]
    \centering
    \caption{Architecture of the DeepONet. Here, lr denotes the learning rate, $\mathrm{\Gamma}$ is the activation function, Dense denotes the feed-forward later, and Conv2D denotes the two-dimensional convolution neural network. Further, k and s denote the kernel and stride.}
    \label{tab:deeponet}
    \begin{tabular}{llllllllll}
    \hline
    Examples & Trunk Net & Branch Net & Scheduler \\
    \hline
    Burgers & Dense:[1024, 512, 512, 256, 128, 128] & Dense:[1, 128, 128, 128, 128] & stepLR, $10^4 @$0.5  \\
    Advection & Dense:[40, 128, 128, 128, 128] & Dense:[1, 128, 128, 128] & stepLR, $10^4 @$0.5  \\
    Allen Cahn & Dense:[257, 512, 512] & Dense:[2, 512, 512, 512] & stepLR, $10^4 @$0.5  \\
    Kuromoto & Dense:[257, 512, 1024, 1024, 512, 512] & Dense:[2, 512, 512, 512, 512, 512] & stepLR, $10^4 @$0.5  \\
    Darcy (rectangular) & [Conv2D(64,k=5,s=2), & Dense:[2, 128, 128, 128, 128] & stepLR, $10^4 @$0.5  \\
    & Conv2D(128,k=5, s=2), & \\
    & Dense(128),Dense(128)] & \\
    Darcy (triangular) & [Conv2D(64,k=5,s=2), & Dense:[2, 128, 128, 128, 128] & stepLR, $10^4 @$0.5  \\
    & Conv2D(128,k=5, s=2), & \\
    & Dense(128),Dense(128)] & \\
    Navier-Stokes & [Conv2D(64,k=5,s=2), & Dense:[3, 128, 128, 128, 128] & stepLR, $10^4 @$0.5  \\
    & Conv2D(128,k=5, s=2), & \\
    & Conv2D(128,k=3,s=1), & \\
    & Dense(128),Dense(128)] & \\
    Nagumo & [Conv2D(65,k=5,s=2), & Dense:[3, 128, 128, 128, 128] & stepLR, $10^4 @$0.5  \\
    & Conv2D(128,k=5, s=2), & \\
    & Conv2D(128,k=3,s=1), & \\
    & Dense(128),Dense(128)] & \\
    \hline
    \end{tabular}
\end{table}

\subsection{Fourier neural operator}
The Gaussian error linear unit (geLU) activation is used in all the examples for training. An initial learning rate of 0.001 and 500 epochs are considered. A batch size of 20 is used. The Adam optimizer with a weight decay $1 \times 10^{-4}$ is used. The network details are given in Table \ref{tab:fno}. 
\begin{table}[ht!]
    \centering
    \caption{Architecture of FNO. StepLR decays the learning rate at every stepsize epoch by $@$ gamma rate. P and Q denote the uplifting and downlifting transformation.}
    \label{tab:fno}
    \begin{tabular}{llllllllll}
    \hline
    Examples & P & Q & Fourier Convolution & Fourier modes & Scheduler \\
    \hline
    Burgers & 64 & [128, 1] & [2, 64, 64, 64, 64] & 16 & stepLR, 50$@$0.5  \\
    Advection & 64 & [128, 1] & [2, 64, 64, 64, 64] & 16 & stepLR, 50$@$0.5  \\
    Allen Cahn & 64 & [128, 1] & [11, 64, 64, 64, 64] & 16 & stepLR, 50$@$0.5  \\
    Kuromoto & 64 & [128, 1] & [11, 64, 64, 64, 64] & 16 & stepLR, 50$@$0.5  \\
    Darcy (rectangular) & 64 & [128, 1] & [3, 64, 64, 64, 64] & 16 & stepLR, 50$@$0.5  \\
    Darcy (triangular) & 64 & [128, 1] & [3, 64, 64, 64, 64] & 16 & stepLR, 50$@$0.5  \\
    Navier-Stokes & 20 & [128, 1] & [12, 20, 20, 20, 20] & 12 & stepLR, 100$@$0.5  \\
    Nagumo & 20 & [128, 1] & [12, 20, 20, 20, 20] & 12 & stepLR, 100$@$0.5  \\
    \hline
    \end{tabular}
\end{table}

\subsection{Wavelet neural operator}
The training of WNO is performed using the Gaussian error linear unit (geLU) activation in all the examples. Further, an initial learning rate of 0.001 and 500 epochs are considered in each example. A batch size of 10 is used in the 1D Burgers and advection equation, and a batch size of 20 is used in other examples. The Adam optimizer with a weight decay $1 \times 10^{-6}$ is used. Table \ref{tab:wno} shows the remaining network details. 
\begin{table}[ht!]
    \centering
    \caption{Architecture of WNO. Here, dbN denotes the Daubechies wavelet basis with N vanishing moments, and biort denotes the Biorthogonal wavelet basis function. P and Q denote the uplifting and downlifting transformation.}
    \label{tab:wno}
    \begin{tabular}{llllllllll}
    \hline
    Examples & P & Q & Wavelet & Wavelet layer & Compression level & Scheduler \\
    \hline
    Burgers & 64 & [128, 1] & [2, 64, 64, 64, 64] & db6 & 8 & stepLR, 50$@$0.5  \\
    Advection & 64 & [128, 1] & [2, 64, 64, 64, 64] & db4 & 2 & stepLR, 50$@$0.5  \\
    Allen Cahn & 64 & [128, 1] & [11, 64, 64, 64, 64] & db6 & 4 & stepLR, 50$@$0.5  \\
    Kuromoto & 64 & [128, 1] & [11, 64, 64, 64, 64] & db8 & 4 & stepLR, 50$@$0.5  \\
    Darcy (rectangular) & 64 & [128, 1] & [3, 64, 64, 64, 64] & biort & 4 & stepLR, 50$@$0.5  \\
    Darcy (triangular) & 64 & [128, 1] & [3, 64, 64, 64, 64] & biort & 4 & stepLR, 50$@$0.5  \\
    Navier-Stokes & 32 & [128, 1] & [12, 32, 32, 32, 32] & biort & 4 & stepLR, 50$@$0.5  \\
    Nagumo & 32 & [128, 1] & [12, 32, 32, 32, 32] & biort & 4 & stepLR, 50$@$0.5  \\
    \hline
    \end{tabular}
\end{table}

\newpage


\begin{thebibliography}{10}

\bibitem{bommasani2021opportunities}
Rishi Bommasani, Drew~A Hudson, Ehsan Adeli, Russ Altman, Simran Arora, Sydney von Arx, Michael~S Bernstein, Jeannette Bohg, Antoine Bosselut, Emma Brunskill, et~al.
\newblock On the opportunities and risks of foundation models.
\newblock {\em arXiv preprint arXiv:2108.07258}, 2021.

\bibitem{veness2021gated}
Joel Veness, Tor Lattimore, David Budden, Avishkar Bhoopchand, Christopher Mattern, Agnieszka Grabska-Barwinska, Eren Sezener, Jianan Wang, Peter Toth, Simon Schmitt, et~al.
\newblock Gated linear networks.
\newblock In {\em Proceedings of the AAAI Conference on Artificial Intelligence}, volume~35, pages 10015--10023, 2021.

\bibitem{thrun1998lifelong}
Sebastian Thrun.
\newblock Lifelong learning algorithms.
\newblock In {\em Learning to learn}, pages 181--209. Springer, 1998.

\bibitem{kenton2019bert}
Jacob Devlin Ming-Wei~Chang Kenton and Lee~Kristina Toutanova.
\newblock Bert: Pre-training of deep bidirectional transformers for language understanding.
\newblock In {\em Proceedings of naacL-HLT}, volume~1, page~2, 2019.

\bibitem{brown2020language}
Tom Brown, Benjamin Mann, Nick Ryder, Melanie Subbiah, Jared~D Kaplan, Prafulla Dhariwal, Arvind Neelakantan, Pranav Shyam, Girish Sastry, Amanda Askell, et~al.
\newblock Language models are few-shot learners.
\newblock {\em Advances in neural information processing systems}, 33:1877--1901, 2020.

\bibitem{radford2021learning}
Alec Radford, Jong~Wook Kim, Chris Hallacy, Aditya Ramesh, Gabriel Goh, Sandhini Agarwal, Girish Sastry, Amanda Askell, Pamela Mishkin, Jack Clark, et~al.
\newblock Learning transferable visual models from natural language supervision.
\newblock In {\em International conference on machine learning}, pages 8748--8763. PMLR, 2021.

\bibitem{jia2021scaling}
Chao Jia, Yinfei Yang, Ye~Xia, Yi-Ting Chen, Zarana Parekh, Hieu Pham, Quoc Le, Yun-Hsuan Sung, Zhen Li, and Tom Duerig.
\newblock Scaling up visual and vision-language representation learning with noisy text supervision.
\newblock In {\em International conference on machine learning}, pages 4904--4916. PMLR, 2021.

\bibitem{chowdhery2022palm}
Aakanksha Chowdhery, Sharan Narang, Jacob Devlin, Maarten Bosma, Gaurav Mishra, Adam Roberts, Paul Barham, Hyung~Won Chung, Charles Sutton, Sebastian Gehrmann, et~al.
\newblock Palm: Scaling language modeling with pathways.
\newblock {\em arXiv preprint arXiv:2204.02311}, 2022.

\bibitem{yuan2023power}
Yang Yuan.
\newblock On the power of foundation models.
\newblock In {\em International Conference on Machine Learning}, pages 40519--40530. PMLR, 2023.

\bibitem{zhou2023comprehensive}
Ce~Zhou, Qian Li, Chen Li, Jun Yu, Yixin Liu, Guangjing Wang, Kai Zhang, Cheng Ji, Qiben Yan, Lifang He, et~al.
\newblock A comprehensive survey on pretrained foundation models: A history from bert to chatgpt.
\newblock {\em arXiv preprint arXiv:2302.09419}, 2023.

\bibitem{guu2020retrieval}
Kelvin Guu, Kenton Lee, Zora Tung, Panupong Pasupat, and Mingwei Chang.
\newblock Retrieval augmented language model pre-training.
\newblock In {\em International conference on machine learning}, pages 3929--3938. PMLR, 2020.

\bibitem{hughes2012finite}
Thomas~JR Hughes.
\newblock {\em The finite element method: linear static and dynamic finite element analysis}.
\newblock Courier Corporation, 2012.

\bibitem{lord2014introduction}
Gabriel~J Lord, Catherine~E Powell, and Tony Shardlow.
\newblock {\em An introduction to computational stochastic PDEs}, volume~50.
\newblock Cambridge University Press, 2014.

\bibitem{moukalled2016finite}
Fadl Moukalled, Luca Mangani, Marwan Darwish, F~Moukalled, L~Mangani, and M~Darwish.
\newblock {\em The finite volume method}.
\newblock Springer, 2016.

\bibitem{lucia2004reduced}
David~J Lucia, Philip~S Beran, and Walter~A Silva.
\newblock Reduced-order modeling: new approaches for computational physics.
\newblock {\em Progress in aerospace sciences}, 40(1-2):51--117, 2004.

\bibitem{peherstorfer2016data}
Benjamin Peherstorfer and Karen Willcox.
\newblock Data-driven operator inference for nonintrusive projection-based model reduction.
\newblock {\em Computer Methods in Applied Mechanics and Engineering}, 306:196--215, 2016.

\bibitem{lassila2014model}
Toni Lassila, Andrea Manzoni, Alfio Quarteroni, and Gianluigi Rozza.
\newblock Model order reduction in fluid dynamics: challenges and perspectives.
\newblock {\em Reduced Order Methods for modeling and computational reduction}, pages 235--273, 2014.

\bibitem{majda2018strategies}
Andrew~J Majda and Di~Qi.
\newblock Strategies for reduced-order models for predicting the statistical responses and uncertainty quantification in complex turbulent dynamical systems.
\newblock {\em SIAM Review}, 60(3):491--549, 2018.

\bibitem{raissi2019physics}
Maziar Raissi, Paris Perdikaris, and George~E Karniadakis.
\newblock Physics-informed neural networks: A deep learning framework for solving forward and inverse problems involving nonlinear partial differential equations.
\newblock {\em Journal of Computational Physics}, 378:686--707, 2019.

\bibitem{sun2020surrogate}
Luning Sun, Han Gao, Shaowu Pan, and Jian-Xun Wang.
\newblock Surrogate modeling for fluid flows based on physics-constrained deep learning without simulation data.
\newblock {\em Computer Methods in Applied Mechanics and Engineering}, 361:112732, 2020.

\bibitem{zhu2019physics}
Yinhao Zhu, Nicholas Zabaras, Phaedon-Stelios Koutsourelakis, and Paris Perdikaris.
\newblock Physics-constrained deep learning for high-dimensional surrogate modeling and uncertainty quantification without labeled data.
\newblock {\em Journal of Computational Physics}, 394:56--81, 2019.

\bibitem{sirignano2018dgm}
Justin Sirignano and Konstantinos Spiliopoulos.
\newblock Dgm: A deep learning algorithm for solving partial differential equations.
\newblock {\em Journal of computational physics}, 375:1339--1364, 2018.

\bibitem{goswami2020transfer}
Somdatta Goswami, Cosmin Anitescu, Souvik Chakraborty, and Timon Rabczuk.
\newblock Transfer learning enhanced physics informed neural network for phase-field modeling of fracture.
\newblock {\em Theoretical and Applied Fracture Mechanics}, 106:102447, 2020.

\bibitem{chakraborty2021transfer}
Souvik Chakraborty.
\newblock Transfer learning based multi-fidelity physics informed deep neural network.
\newblock {\em Journal of Computational Physics}, 426:109942, 2021.

\bibitem{li2020fourier}
Zongyi Li, Nikola Kovachki, Kamyar Azizzadenesheli, Burigede Liu, Kaushik Bhattacharya, Andrew Stuart, and Anima Anandkumar.
\newblock Fourier neural operator for parametric partial differential equations.
\newblock {\em arXiv preprint arXiv:2010.08895}, 2020.

\bibitem{li2020neural}
Zongyi Li, Nikola Kovachki, Kamyar Azizzadenesheli, Burigede Liu, Kaushik Bhattacharya, Andrew Stuart, and Anima Anandkumar.
\newblock Neural operator: Graph kernel network for partial differential equations.
\newblock {\em arXiv preprint arXiv:2003.03485}, 2020.

\bibitem{gupta2021multiwavelet}
Gaurav Gupta, Xiongye Xiao, and Paul Bogdan.
\newblock Multiwavelet-based operator learning for differential equations.
\newblock {\em Advances in Neural Information Processing Systems}, 34, 2021.

\bibitem{lu2021learning}
Lu~Lu, Pengzhan Jin, Guofei Pang, Zhongqiang Zhang, and George~Em Karniadakis.
\newblock Learning nonlinear operators via deeponet based on the universal approximation theorem of operators.
\newblock {\em Nature Machine Intelligence}, 3(3):218--229, 2021.

\bibitem{wang2021learning}
Sifan Wang, Hanwen Wang, and Paris Perdikaris.
\newblock Learning the solution operator of parametric partial differential equations with physics-informed deeponets.
\newblock {\em Science advances}, 7(40):eabi8605, 2021.

\bibitem{tripura2023wavelet}
Tapas Tripura and Souvik Chakraborty.
\newblock Wavelet neural operator for solving parametric partial differential equations in computational mechanics problems.
\newblock {\em Computer Methods in Applied Mechanics and Engineering}, 404:115783, 2023.

\bibitem{tripura2023physics}
Tapas Tripura, Souvik Chakraborty, et~al.
\newblock Physics informed wno.
\newblock {\em arXiv preprint arXiv:2302.05925}, 2023.

\bibitem{yosida2012functional}
K{\"o}saku Yosida.
\newblock {\em Functional analysis}.
\newblock Springer Science \& Business Media, 2012.

\bibitem{garg2023vb}
Shailesh Garg and Souvik Chakraborty.
\newblock Vb-deeponet: A bayesian operator learning framework for uncertainty quantification.
\newblock {\em Engineering Applications of Artificial Intelligence}, 118:105685, 2023.

\bibitem{chen1995universal}
Tianping Chen and Hong Chen.
\newblock Universal approximation to nonlinear operators by neural networks with arbitrary activation functions and its application to dynamical systems.
\newblock {\em IEEE Transactions on Neural Networks}, 6(4):911--917, 1995.

\bibitem{back2002universal}
Andrew~D Back and Tianping Chen.
\newblock Universal approximation of multiple nonlinear operators by neural networks.
\newblock {\em Neural Computation}, 14(11):2561--2566, 2002.

\bibitem{navaneeth2022koopman}
N~Navaneeth and Souvik Chakraborty.
\newblock Koopman operator for time-dependent reliability analysis.
\newblock {\em Probabilistic Engineering Mechanics}, 70:103372, 2022.

\bibitem{tripura2023elastography}
Tapas Tripura, Abhilash Awasthi, Sitikantha Roy, and Souvik Chakraborty.
\newblock A wavelet neural operator based elastography for localization and quantification of tumors.
\newblock {\em Computer Methods and Programs in Biomedicine}, 232:107436, 2023.

\bibitem{li2021physics}
Zongyi Li, Hongkai Zheng, Nikola Kovachki, David Jin, Haoxuan Chen, Burigede Liu, Kamyar Azizzadenesheli, and Anima Anandkumar.
\newblock Physics-informed neural operator for learning partial differential equations.
\newblock {\em arXiv preprint arXiv:2111.03794}, 2021.

\bibitem{kadri2016operator}
Hachem Kadri, Emmanuel Duflos, Philippe Preux, St{\'e}phane Canu, Alain Rakotomamonjy, and Julien Audiffren.
\newblock Operator-valued kernels for learning from functional response data.
\newblock {\em Journal of Machine Learning Research}, 2016.

\bibitem{griebel2017reproducing}
Michael Griebel and Christian Rieger.
\newblock Reproducing kernel hilbert spaces for parametric partial differential equations.
\newblock {\em SIAM/ASA Journal on Uncertainty Quantification}, 5(1):111--137, 2017.

\bibitem{wang2020combinatorial}
Jianan Wang, Eren Sezener, David Budden, Marcus Hutter, and Joel Veness.
\newblock A combinatorial perspective on transfer learning.
\newblock {\em Advances in Neural Information Processing Systems}, 33:918--929, 2020.

\bibitem{misra2019mish}
Diganta Misra.
\newblock Mish: A self regularized non-monotonic activation function.
\newblock {\em arXiv preprint arXiv:1908.08681}, 2019.

\bibitem{jeeveswaran2023birt}
Kishaan Jeeveswaran, Prashant Bhat, Bahram Zonooz, and Elahe Arani.
\newblock Birt: Bio-inspired replay in vision transformers for continual learning.
\newblock {\em arXiv preprint arXiv:2305.04769}, 2023.

\bibitem{daubechies1992ten}
Ingrid Daubechies.
\newblock {\em Ten lectures on wavelets}.
\newblock SIAM, 1992.

\bibitem{lu2022comprehensive}
Lu~Lu, Xuhui Meng, Shengze Cai, Zhiping Mao, Somdatta Goswami, Zhongqiang Zhang, and George~Em Karniadakis.
\newblock A comprehensive and fair comparison of two neural operators (with practical extensions) based on fair data.
\newblock {\em Computer Methods in Applied Mechanics and Engineering}, 393:114778, 2022.

\bibitem{geneva2020modeling}
Nicholas Geneva and Nicholas Zabaras.
\newblock Modeling the dynamics of pde systems with physics-constrained deep auto-regressive networks.
\newblock {\em Journal of Computational Physics}, 403:109056, 2020.

\end{thebibliography}

\end{document}